\documentclass{article}

\usepackage{microtype}
\usepackage{graphicx}
\usepackage{subfigure}
\usepackage{booktabs} 

\usepackage{hyperref}

\usepackage[accepted]{icml2021}

\usepackage{amsmath}
\usepackage{amssymb}
\usepackage{amsthm}
\usepackage{mathrsfs}
\usepackage{bm}

\newtheorem{theorem}{Theorem}
\newtheorem{proposition}{Proposition}
\DeclareMathOperator{\tr}{tr}

\newcommand{\bb}[1]{\mathbf{#1}}
\newcommand{\bgreek}[1]{\boldsymbol{#1}}
\newcommand{\be}{\bb{e}}
\newcommand{\bh}{\bb{h}}
\newcommand{\bs}{\bb{s}}
\newcommand{\bt}{\bb{t}}

\newcommand{\bx}{\bb{x}}
\newcommand{\by}{\bb{y}}
\newcommand{\bz}{\bb{z}}
\newcommand{\bM}{\bb{M}}
\newcommand{\bQ}{\bb{Q}}
\newcommand{\bS}{\bb{S}}
\newcommand{\bW}{\bb{W}}
\newcommand{\bg}{\bm{g}}
\newcommand{\bbf}{\bm{f}}

\newcommand{\btheta}{\bgreek{\theta}}

\DeclareMathOperator*{\argmax}{arg\,max}
\newcommand{\fpartial}[2]{\frac{\partial{#1}}{\partial{#2}}}

\icmltitlerunning{Jacobian Determinant of Normalizing Flows}

\begin{document}

\twocolumn[
\icmltitle{Jacobian Determinant of Normalizing Flows}

\begin{icmlauthorlist}
\icmlauthor{Huadong Liao}
\icmlauthor{Jiawei He}
\end{icmlauthorlist}

\icmlcorrespondingauthor{Huadong Liao}{naturomics.liao@gmail.com}

\icmlkeywords{jacobian determinant, normalizing flows, stability, generative flows, density estimation}

\vskip 0.3in
]

\printAffiliationsAndNotice{}  

\begin{abstract}
Normalizing flows learn a diffeomorphic mapping between the target and base distribution, while the Jacobian determinant of that
mapping forms another real-valued function. In this paper, we show that the Jacobian determinant mapping is unique for the given
distributions, hence the likelihood objective of flows has a unique global optimum. In particular, the likelihood for a class
of flows is explicitly expressed by the eigenvalues of the auto-correlation matrix of individual data point, and independent of the
parameterization of neural network, which provides a theoretical optimal value of likelihood objective and relates to probabilistic PCA.
Additionally, Jacobian determinant is a measure of local volume change and is maximized when MLE is used for optimization.
To stabilize normalizing flows training, it is required to maintain a balance between the expansiveness and contraction of volume,
meaning Lipschitz constraint on the diffeomorphic mapping and its inverse. With these theoretical results, several principles of
designing normalizing flow were proposed. And numerical experiments on high-dimensional datasets (such as CelebA-HQ $1024^2$)
were conducted to show the improved stability of training.
\end{abstract}

\section{Introduction}
\label{introduction}

\begin{figure}
  \centering
  \includegraphics[width=0.46\textwidth]{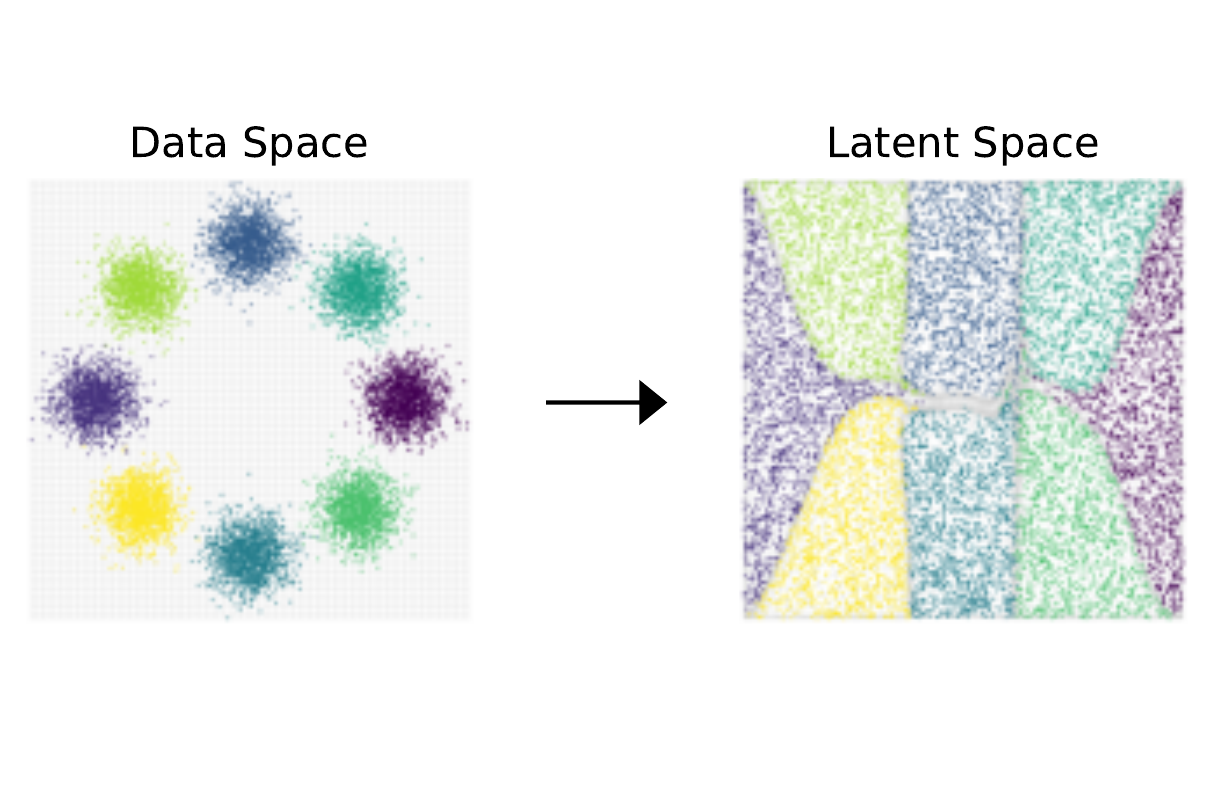}
  \vspace{-1.2cm}
  \caption{Normalizing flows are diffeomorphic maps and able to provide topology-preserving latent representations,
  where the latent is uniformly distributed on the same domain as the data space.}
  \label{fig:latent}
  \vspace{-0.1cm}
\end{figure}

Density estimation is a core paradigm in machine learning that aims to learn the underlying
representation of data distribution, mathematically, to estimate an unobservable probability
density $p_{\bm{X}}$ based on an i.i.d. dataset $\bm{D}=\{\bx^{(n)}\}_{n=1}^N$ drawn from that distribution.
This task is challenging in real world applications, since feature $\bx$ is usually high-dimensional and complex,
thus one can not parameterize $p_{\bm{X}}$ directly.
Recently, an attractive solution to this task, called normalizing flows, has gained great popularity for its efficient
and exact evaluation on inference and sampling, and useful latent representation for downstream tasks (Fig.~\ref{fig:latent}). 
More specifically, normalizing flows are derived from the change of variables theorem,
estimating probability density by leveraging a sequence of diffeomorphic mappings
$\bg_{\btheta}=\bg_1\circ\bg_2\circ\cdots\circ\bg_L: \mathbb{R}^d\to\mathbb{R}^d$ with inverse $\bbf=\bg^{-1}$.
Here $\btheta=\{\btheta_1,\btheta_2,\cdots,\btheta_L\}$ is the parameters of $\bg$ and the latent representation $\bz=\bbf(\bx)$
is assumed to follow a distribution with known form probability density $q_{\bm{Z}}$. Given by the theorem, the likelihood
for a point $\bx$ can be obtained by
\begin{align}
  p(\bx) &= q(\bbf(\bx))|\det(J_{\bbf}(\bx))| \label{eq:changeofvar}\\
         &= q(\bbf(\bx)) \prod_{l=1}^{L}\left|\det(J_{\bbf_l}(\bh_l))\right|, \notag
\end{align}
where $J_{\bbf}(\bx):={\partial\bbf(\bx)}/{\partial\bx}$ is the Jacobian of $\bbf$ w.r.t $\bx$,
$\det(\cdot)$ denotes the determinant, and $\bh_l$ denotes the output of intermediate mapping $\bg_l$,
with $\bh_1 = \bx$ and $\bh_{L} = \bg_L(\bz)$. Given the observed dataset $\bm{D}$, the parameters
$\btheta$ can be learned using statistical technique such as maximum likelihood estimation (MLE):
\begin{equation}
\begin{aligned}
  \btheta^{*} = \argmax_{\btheta}{\mathcal{L}},\ \mathcal{L}(\btheta;\bm{D}) = \sum_{n=1}^N \log p_{\btheta}(\bx^{(n)}).
  \label{eq:mle}
\end{aligned}
\end{equation}

Theoretically, normalizing flow is powerful to learn probability distribution with arbitrary complexity, 
supposing the mapping $\bg$ is expressive enough. But in practice, there is an obstacle posed by the computation of determinant term,
since it has a cubic cost in the dimension of the Jacobian matrix.

To address this computational challenge, at least three strategies have been investigated in the machine learning community.
A first approach involves the application of the matrix determinant lemma~\citep{rezende2015NF,vdberg2018sylvester},
which converts the calculation of Jacobian determinant into the determinant of a lower rank matrix,
thus the cost is reduced to the cube of the dimension of the lower rank matrix.
A second approach involves the stochastic approximation of log-determinant. Example of this approach is to expand the log-determinant
into a power series in terms of the trace of power of Jacobian~\citep{benhrmann2019,grathwohl2019ffjord, chen2019residualflows}.
A third approach involves a basic property in linear algebra that the determinant for matrices in special form is cheap to calculate,
e.g., the determinant of triangular or diagonal matrix is simply the product of its diagonal terms, in which case the cost is linear
in dimensionality. Partition-based flows, which are further divided into flow-based~\citep{dinh2014nice,dinh2016density}
and autoregressive models~\citep{kingma2016improved,papamakarios2017masked},
utilize this property by splitting the input of model into parts and constructing ordered
dependencies between parts (i.e. the transform on the $i$-th part only depends on parts 1 to $i$),
enforcing Jacobian matrix to be triangular. This family is popular due to (1) its exact likelihood, (2) computational tractability,
and (3) analytic inverse, while the others lack one or two of these features.

The consequence of restricting determinant is limited expressivity of flows. To make up for this limitation,
a main branch of research on flows is to construct more
powerful diffeomorphic mapping under the above strategies. Remarkable work within this line includes Glow~\citep{kingma2018glow},
Flow++~\citep{ho2019flowpp}, Augmented Flows~\citep{huang2020augmented}, ResFlows~\citep{chen2019residualflows}, among others.
Meanwhile, another branch tries to find out whether these restricted flows are universal for arbitrary distributions in
theory\citep{kong2020expressive,teshima2020coupling,koehler2020rep,huang2020solving}.
Moreover, optimal transport theory was introduced to provide different convergence properties
from MLE~\citep{zhang2018monge,yang2020potential,onken2020ot}.

Despite these tremendous advances, the property of Jacobian determinant was not well studied in the context of normalizing flows.
The Jacobian determinant can be seen as another mapping different from the diffeomorphism, and is a part of optimization objective.
From this view, it is natural to ask (\romannumeral1) if the determinant mapping given by flows is unique, (\romannumeral2) what the
relation is between its continuity and convergence, and (\romannumeral3) how it affects the performance of flows.
In this work, we focus on these questions, aiming to provide a better understanding of normalizing flows.

Our contributions are summarized as follows.
\begin{itemize}
  \item We show the Jacobian determinant mapping of flows is unique for two given distributions, when there are multiple equivalent
  diffeomorphisms corresponding to that mapping. In particular, we show the determinant mapping for a class of flows has a closed form,
  thus a theoretical global optimum of likelihood objective is available. For such flows, the relation to PPCA is further built.
  \item We present there is an equilibrium between the expansion and contraction of volume. This balance requires bounded determinant,
  and Lipschitz constraint on the diffeomorphism and its inverse to ensure convergence.
  \item Based on our theoretical results, we propose a new flow and demonstrate its improved stability on high-resolution natural images
  (CelebA HQ 1024$\times$1024). In addition, various experiments were performed to explain the dynamics of normalizing flows.
\end{itemize}

\section{Properties of Jacobian Determinant}
\label{sec:properties}

In this section, we begin by discussing the existence and uniqueness of the solution of normalizing flows, specifically showing
the closed form of likelihood objective for a subset of flows and the relation of those flows to Probabilistic Principal Component
Analysis(PPCA; \citealt{Tipping99probabilisticpca}). We then analyze the optimization behaviour of normalizing flows,
and give out the conditions on robust training.

\subsection{Existence and Uniqueness}
\subsubsection{general flow}
\label{sec:genflow}
Given a measure space $(\Omega,\mathcal{F}, \mu)$, a measurable space $(\Omega^\prime, \mathcal{F}^\prime)$,
and a measurable mapping $\bg:\Omega\to\Omega^\prime$, one can define a push-forward meansure
as $\bg_{\ast\mu}(U)=\mu(\bg^{-1}(U))$, for all $U\in\mathcal{F}^\prime$. For the problem of representation learning or generative
modelling (e.g., normalizing flows), one can interpret $\Omega$ as a latent spcae, given a set of samples from a measured "data" space
$(\Omega^\prime,\mathcal{F}^\prime,\nu)$, the task is to find a function $\bg$ such that $\bg_{\ast\mu}=\nu$.
The existence of $\bg$ can be guaranteed by the Radon-Nikodym theorem~\citep{rudin1987real} when extra conditions satisfied:
\begin{theorem}
\textnormal{(Radon-Nikodym)}
Let $\mu$ and $\nu$ be two $\sigma$-finite measures on the same measurable space $(\Omega, \mathcal{F})$,
if $\nu$ is absolutely continuous with respect to $\mu$,
then (a) there is an $\Omega$-measurable function $\tau: \Omega\to [0,\infty)$, such that $\nu(U)=\int_U \tau d\mu$,
for all $U\in\mathcal{F}$; (b) such function is unique upto a.e. equality w.r.t. $\mu$.
\label{thm:radon}
\end{theorem}
The function $\tau$ is called Radon-Nikodym derivative, and commonly written as $\tau=\frac{d\nu}{d\mu}$.
Mapping $\bg$ exists and is any function satisfying $\bg_{\ast\mu}=\nu$ and $|\det(J_{\bg^{-1}})|=\tau$.
If we further restrict $\mu$ and $\nu$ to probability measures (i.e., $\mu(\Omega)=\nu(\Omega)=1$), which is a basic
assumption in normalizing flows literature, then $\tau$ corresponds to the determinant term
of Eq.~\eqref{eq:changeofvar}, or the density ratio ${p(\bx)}/{q(\bg^{-1}(\bx))}$ between target and base distribution.
Using probability terminology, Theorem~\ref{thm:radon} implies that if two random vectors $\bm{X}$ and
$\bm{Z}$ are on the same sample space,
and there is a bijective and absolutely continuous mapping $\bg: \bm{Z}\to\bm{X}$ (whose inverse is $\bbf=\bg^{-1}$),
then \emph{there is a unique mapping} $\tau:\bx\to [0,\infty)$ \emph{such that}
$\tau=\frac{p(\bx)}{q(\bbf(\bx))}=|\det(J_{\bbf})|$. Note the uniqueness of $\tau$ does not imply the uniqueness of $\bg$,
e.g., $\tau$ is equivalent between $\bg=\bbf^{-1}$ and $\bg^\prime=(\bQ\bbf)^{-1}$, for $\bQ$ being an arbitrary orthogonal matrix.

In the rest of this paper, if not specified, we assume $\bz$ is uniformly distributed in $(0,1)^d$ (i.e., $\mu$ is a Lebesgue measure).
Because if $q_{\bm{Z}}$ is not uniform, we can always find an additional mapping $\bbf^\prime$ to transform $\bz$ to a uniform and let its
Jacobian determinant be exactly the density $q_{\bm{Z}}$, so the problem does not change but has a new transform
$\bg\leftarrow\bg\circ {\bbf^{\prime}}^{-1},\ \bbf\leftarrow\bbf^{\prime}\circ\bbf$. With this setting, Eq.~\eqref{eq:changeofvar} is
simplified to
\begin{equation}
  p(\bx)=\tau:\bx\mapsto |\det(J_{\bbf}(\bx))|.
  \label{eq:jac}
\end{equation}
We refer Eq.~\eqref{eq:jac} as Jacobian determinant mapping, $\bbf$ and $\bg$ as diffeomorphic mappings in the next.

\subsubsection{quasi-linear flow}
We have shown the uniqueness of Jacobian determinant mapping by Radon-Nikodym theorem. Here we further present the detailed form of
$\tau$ when additional constraint is applied to the diffeomorphic mapping.

Consider normalizing flows in the following special form:
\begin{equation}
\begin{aligned}
  \bbf(\bx) = \bW(\bx)&\bx + \bb{b}(\bx), \\
  \text{s.t. } \bbf \sim\mathcal{N}(\bm{0},\bm{I}),\ \det(J_{\bbf})&=\det(\bW), \forall \bx\in\bm{X},
\end{aligned}
\end{equation}
where $\bW(\bx)\in \mathbb{R}^{d\times d}$ and $\bb{b}(\bx)\in\mathbb{R}^d$ are parameterized by or independent of $\bx$. To simplify
the analysis, we omit the bias term $\bb{b}(\bx)$ in the following, as it can be treated as part of $\bW(\bx)$ by adding some constant
dimensions to $\bx$ (or assuming $\bx$ is zero-centered thus bias term is zero).
A class of previously proposed flows can be rewritten in this form, including linear mapping~\citep{kingma2018glow},
affine/additive coupling layers~\citep{dinh2014nice,dinh2016density},
dynamic linear layers~\citep{liao2019dlf}, affine autoregressive flows~\citep{kingma2016improved,papamakarios2017masked},
and their compositions. We call them Quasi-Linear Flow (QLF), since the formula is similar to a linear function,
and it does degenerate to a conventional linear mapping when $\bW$ and $\bb{b}$ are independent of $\bx$.

For QLF, we have $\tau=\mathcal{N}(\bbf;\bm{0},\bm{I})\det(\bW)$, and the corresponding log-likelihood:
\begin{equation}
 \mathcal{L} = \mathbb{E}_{p_{\bm{X}}}\left[-\frac{1}{2}\left\{d\log 2\pi\!+\!\tr(\bM\bS)\!+\!\log|\det(\bM^{-1})|\right\}\right],
 \label{eq:loglikelihood}
\end{equation}
where $\bM=\bW(\bx)^T\bW(\bx)$, and $\bS=\bx\bx^T$ is the auto-correlation matrix of individual data point.
The above log-likelihood is maximized when $\bW(\bx)=\bb{U}\bb{\Lambda}^{-1/2}\bb{V}^T$,
in which $\bb{U}$ is an arbitrary $d\times d$ orthogonal matrix, and $\bb{V}$ is also a $d\times d$ matrix whose columns are the eigenvectors
of $\bS$, with $\bb{\Lambda}=\text{diag}(\lambda_1,\lambda_2,\cdots,\lambda_d)$ the corresponding diagonal matrix of eigenvalues.
Substituting the results into $\mathcal{L}$, the global maximum of the log-likelihood is uniquely given by
\begin{equation}
 \mathcal{L}_\text{max} = \mathbb{E}_{p_{\bm{X}}}\left[-\frac{1}{2}\left\{d\log(2\pi) + d + \sum_{i=1}^{d}\log\lambda_i\right\}\right].
 \label{eq:maxll}
\end{equation}
The $\mathcal{L}_\text{max}$ is independent of the parameters of flow, and available based on the observed dataset.
This provides a theoretical optimal value of likelihood objective.

\paragraph{From PPCA to QLF.}
QLF can be treated as a nonlinear extension of Probabilistic PCA,
where PPCA is the case when $\bW$ and $\bb{b}$ are independent of $\bx$,
i.e., globally shared over all points $\bx\sim p_{\bm{X}}$. As a generalization, QLF describes flows
that can be approximated by a set of linear functions locally, and is allowed to stack multi-layers to get a highly complex model.
For PPCA, the expectation in Eq.~\eqref{eq:loglikelihood} can be moved into the trace operation $\tr(\cdot)$, therefore
$\mathcal{L}_\text{max}$ is given by the eigenvalues of covariance matrix of dataset~\citep{Tipping99probabilisticpca},
instead of auto-correlation matrix of individual sample.
PPCA is typically discussed in the context of dimensionality reduction,
where the eigenvalues are descending-ordered and the corresponding first $r (r\le d)$
eigenvectors are called principal axes. Normalizing flows does not have the concept of "principal axes",
but it inherits many properties of PPCA, e.g., the learned model captures
the variance in different direction of data space rather than high level concept such as object semantic, which explains
the observation in previous study (Appendix D in~\citealt{dinh2016density}).

\subsection{Equilibrium between Expansion and Contraction}
\label{sec:balance}
As far, we have theoretical guarantees for the existence and uniqueness of flows.
However, more conditions are needed for its convergence,
particularly when the optimization is calculated using a machine with limited numerical accuracy.
For example, the phenomenon of training instability was observed in previous work~\citep{dupont2019augmented,meng2020gaussianization}
and practical applications.~\footnote{For example, https://github.com/openai/glow/issues/40.}
In this section, we provide an intuitive understanding of optimization behaviour of flows.

As we know, the Jacobian determinant is a measure of local volume change given by a differentiable function,
and is also approximately proportional to the variance change for points in a small region.
Note that the objective of normalizing flows is to maximize likelihood (Eq.~\eqref{eq:mle}), thus plays an expansive effect
on volume. If no constraint was applied to the diffeomorphic mapping $\bbf$,
the determinant $\tau$ will constantly increase as the training iterations update, as well as the variance of output of hidden layers,
finally leads to the problem of gradient exploding or vanishing.
Now consider the basic assumption $\bbf\sim \mathcal{U}_d(0,1)$ aforementioned in Sec.~\ref{sec:genflow}, and let $\bbf=\bbf_a\circ\bbf_b$
in which $\bbf_a$ is the last layer of $\bbf$. The last layer $\bbf_a$ is usually chosen from functions whose range are explicitly
bounded in $(0,1)^d$, e.g., the cumulative distribution function (CDF) of gaussian distribution,
thus $\bbf_b\sim \mathcal{N}(\bm{0},\bm{I})$ and
\begin{equation}
  \log\tau\propto \log|\det J_{\bbf_b}|-\frac{1}{2}\bbf_b^T\bbf_b.
\label{eq:logtau}
\end{equation}
To maximize $\log\tau$, the determinant term in Eq.~\eqref{eq:logtau} still plays an expansiveness effect, while the other term
plays an anti-effect by encouraging $\bbf_b$ to be distributed around the origin.
This subtle balance is strong enough to guarantee the stability of flows with shallow layers, but as the more layers are stacked, 
the greater the challenge is to transfer the contraction effect to the early layer by backpropagation.

\begin{figure}[t]
  \centering
  \subfigure[ImageNet 32$\times$32]{
    \includegraphics[width=0.45\linewidth]{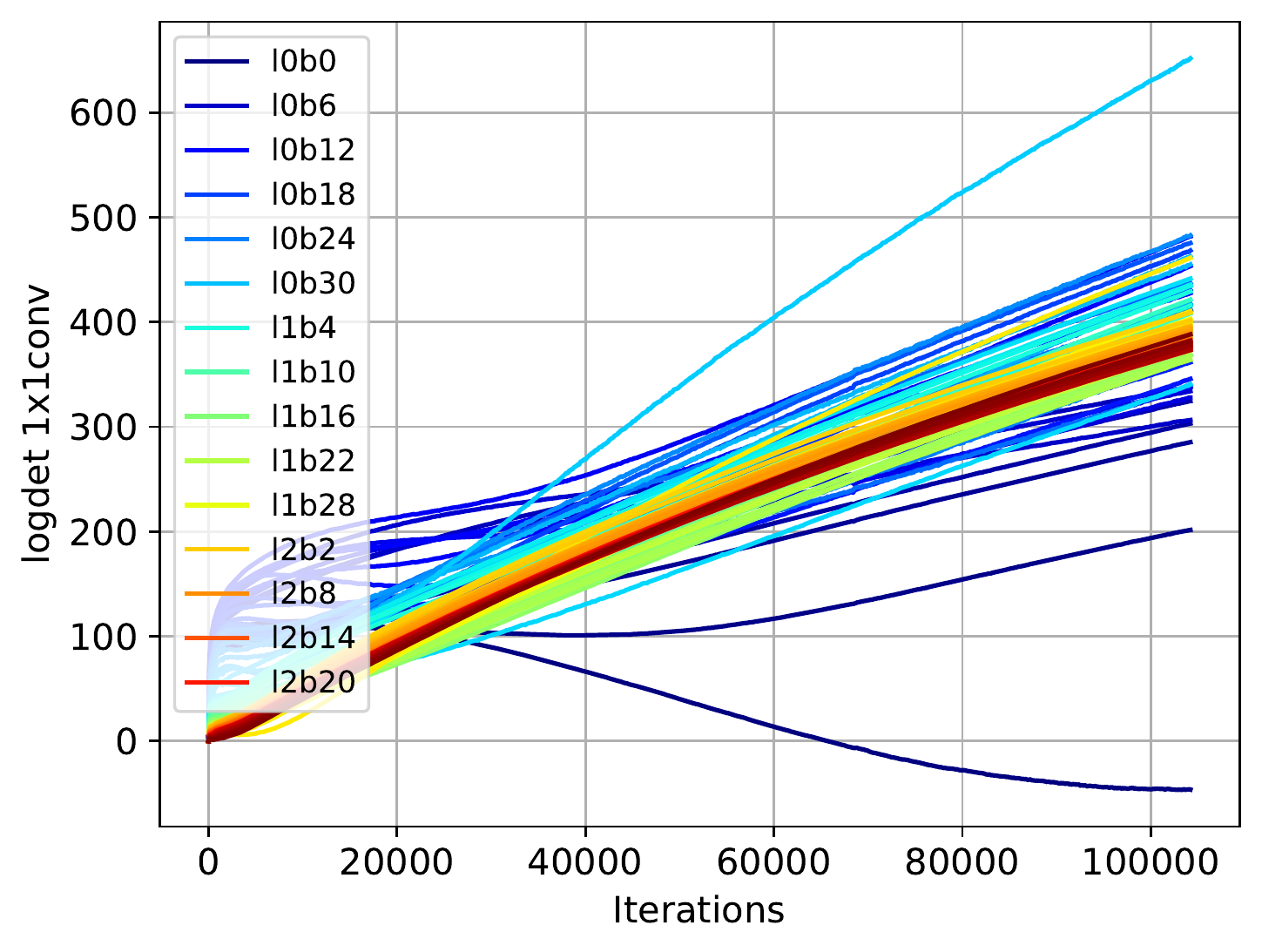}
  }
  \subfigure[CelebA 256$\times$256]{
    \includegraphics[width=0.46\linewidth]{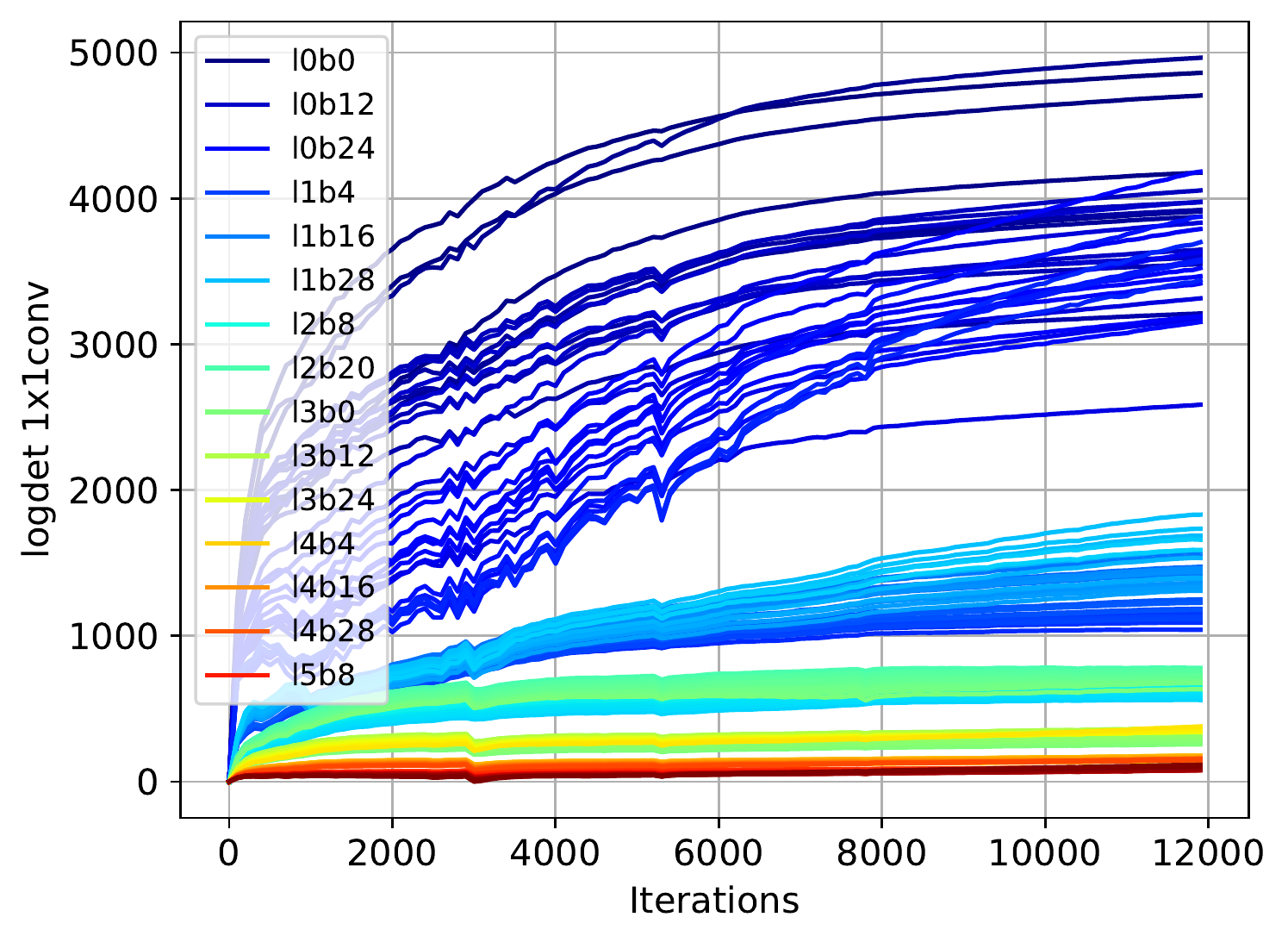}
  }
  \caption{Training curves of log-determinant for invertible 1$\times$1 convolution in each layer of flow. The value is increasing over
  iterations as it is maximized by the objective, where the trend is particularly obvious for earlier layers,
  and the scale is proportional to the dimension of input features ((a) vs. (b)).
  This result is obtained by training Glow with 256 (vs. 512 in official implementation) hidden units.
  The legend denotes the order of level and block for each layer.}
  \label{fig:logdet_conv}
\end{figure}

Fig.\ref{fig:logdet_conv} shows a counterexample based on Glow~\citep{kingma2018glow}.
The invertible 1$\times$1 convolution introduced by this study is a linear mapping for permuting the dimensions between two coupling layers,
and has shown improvement on performance.
However, it is also found unstable after increasing the depth of the model and/or the dimension of input features.
The determinant contributed by this module is given by the determinant of convolution weight and independent of input data.
The only way to prevent the weight turning to infinity is the contraction regularization by backpropagation, which becomes weak for early
layers in a deep model. In a subsequent study by~\citet{liao2019dlf},
$L_2$ regularization was applied to the weight to penalize the unreasonable increase of determinant, and improvement on
stability was observed. In the next, we will explain this phenomenon by analyzing the boundedness of Jacobian determinant mapping.

\subsection{Conditions on Bounded Gradient}
Suppose $\bbf$ is $K$-Lipschitz continuous. By the differentiability of $\bbf$, we have $\|J_{\bbf}(\bx)\|\le K, \forall\bx\in \bm{X}$.
Furthermore, the Lipschitz constant $K$ is related to Jacobian determinant by Hadamard's inequality:
\begin{equation}
  |\det J_{\bbf}(\bx)|\le\prod_{i=1}^d\|J_{\bbf}(\bx)\be_i\|\le \|J_{\bbf}(\bx)\|^d\le K^d,
  \label{eq:bound}
\end{equation}
where $\be_i$ is a unit eigenvector of $J_{\bbf}(\bx)$. The Eq.~\eqref{eq:bound} builds a connection between the boundedness of
$\tau$ and the Lipschitzness of $\bbf$, in which the $d$-power of Lipschitz constant of $\bbf$ is an upper bound of
$\tau$.~\footnote{Commonly only positive determinant is considered thus also bounded below by zero.}
For a fixed $K$, $\tau$ has an exact maximum, which is not what we want as our goal is to maximize $log\tau$.
The question is, whether it is reasonable to let $K$ be finite and $\tau$ bounded.

We answer this question by analyzing the derivatives of log-likelihood.
Since $\nabla_{\theta}\mathbb{E}[\log p(\bx;\btheta)]=\mathbb{E}[\nabla_{\theta}\log p(\bx;\btheta)]$,
let us consider this score function at a single point, and decompose it as follows:
\begin{equation}
  \fpartial{\log p(\bx;\btheta)}{\btheta} = \{\fpartial{\log p(\bx;\btheta)}{\btheta_l}, l=1,2,\dots,L\}.
\end{equation}
To simplify notations, we denote $\nabla_{\btheta_l}$ and $\nabla_{\bh_l}$ the derivatives of $\log p(\bx)$ w.r.t.
$\btheta_l$ and $\bh_l$, respectively. Given by the chain rule, we have the following recursive formulas:
\begin{align}
  \nabla_{\bh_l} = J_{\bbf_l}^T(\bh_l)\nabla_{\bh_{l+1}} + \fpartial{\log|\det J_{\bbf_l}|}{\bh_l}, \label{eq:gradh} \\
  \nabla_{\btheta_l} = \fpartial{\bh_{l+1}}{\btheta_l^T}\nabla_{\bh_{l+1}} + \fpartial{\log|\det J_{\bbf_l}|}{\btheta_l},
  \label{eq:gradtheta}
\end{align}
for $1\le l \le L$ and initial condition $\nabla_{\bh_{L+1}}=\bm{0}$ a zero vector.
From Eq.~\eqref{eq:gradh} and~\eqref{eq:gradtheta}, by the triangle inequality and sub-multiplicativity property of norms, we have
\begin{align}
  \|\nabla_{\btheta_l}\| &\le \|\fpartial{\bh_{l+1}}{\btheta_l}\| \|\nabla_{\bh_{l+1}}\| + \|\fpartial{\log|\det J_{\bbf_l}|}{\btheta_l}\| \\
                         &\le \|\fpartial{\bh_{l+1}}{\btheta_l}\|\cdot\prod_{j=l+1}^{L}\|J_{\bbf_j}\| + o, \label{eq:grad_theta}
\end{align}
in which $o=o(\|\fpartial{\bh_i}{\btheta_i}\|,\|\fpartial{\log|\det J_{\bbf_i}|}{\btheta_i}\|, \|\fpartial{\log|\det J_{\bbf_i}|}{\bh_i}\|)$
($i\ge l$) is a non-negative residual term, and the equality holds under certain conditions. Therefore, we have the following statement.
\begin{proposition}
\textnormal{(Bounded gradient requires bounded Jacobian determinant)}
For $1\le l \le L$, if $\|\nabla_{\btheta_l}\|$ is bounded from above by a constant $C>0$, then there exists a constant
$K>0$ such that $\sum_{j=l+1}^{L}\log|\det J_{f_j}|\le d\log K$. Furthermore, each component $|\det J_{\bbf_l}|$ is bounded.
\label{proposition:logdet-bound}
\end{proposition}
The proof is straightforward by Eq.~\eqref{eq:bound} and~\eqref{eq:grad_theta}. If $\prod_j\|J_{\bbf_j}\|$ is unbounded,
$\|\nabla_{\btheta_l}\|$ is unbounded, and so as $\sum_j{\log\det(J_{\bbf_j})}$. See the Appendix for detailed derivation.

By Proposition~\ref{proposition:logdet-bound}, for a fixed $C$, when the depth of flow $L$ and the dimension of feature $d$ increase,
the Lipschitz constant $K_{\bbf_i}$ for every intermediate mapping $\bbf_i$ should decrease to ensure the gradients of early
layers being bounded. This explains the difference between Fig~\ref{fig:logdet_conv}(a) and (b), where the log-determinant
of higher dimension dataset is significantly larger than the one of lower dimension dataset, and the problem of gradient instability
is more common in models for high-dimensional dataset. On the other hand, $|\det J_{\bbf}|>0$ follows
from the invertibility of $\bbf$, so $\det J_{\bbf^{-1}}<K^\prime$ and the inverse $\bbf^{-1}$ is also Lipschitz continuous.
The Lipschitzness on $\bbf$ and its inverse guarantees the volume expansion and contraction balance discussed in Sec.~\ref{sec:balance}.

\section{Principles of Designing Normalizing Flow}
\label{sec:principles}
In this section, we further discuss normalizing flows from practical aspects.

\begin{figure*}[t]
  \centering
  \setlength{\tabcolsep}{1.5pt}
  \begin{tabular}{ccc}
    \raisebox{0.01\height}{\includegraphics[width=0.102\textwidth]{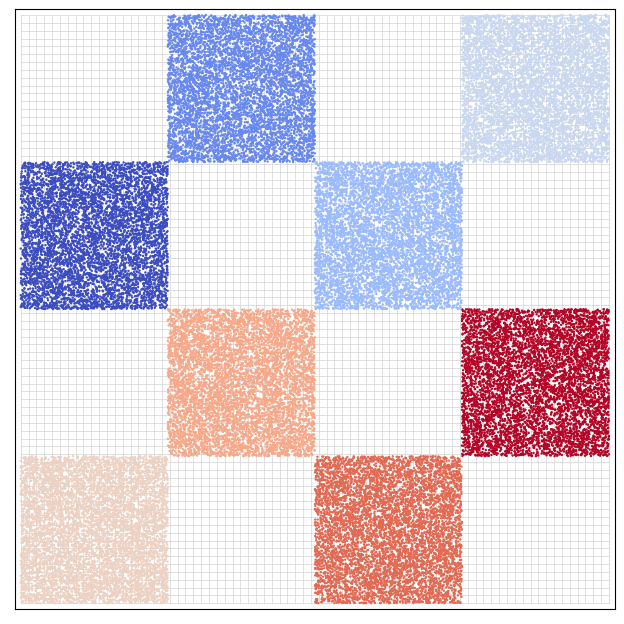}} &
    \includegraphics[width=0.6\textwidth]{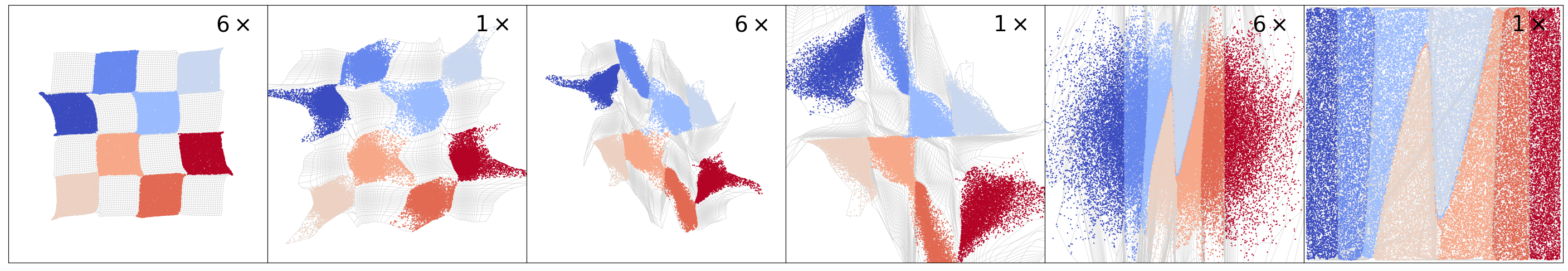} &
    \raisebox{0.02\height}{\includegraphics[width=0.102\textwidth]{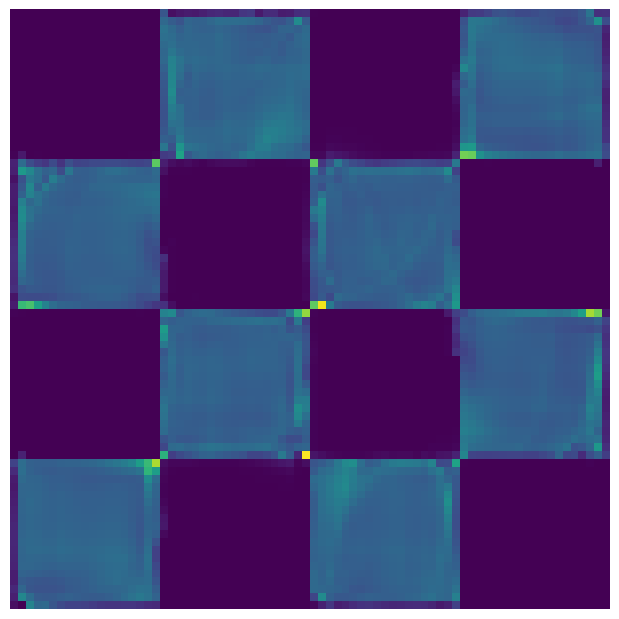}} \\
    \raisebox{0.01\height}{\includegraphics[width=0.102\textwidth]{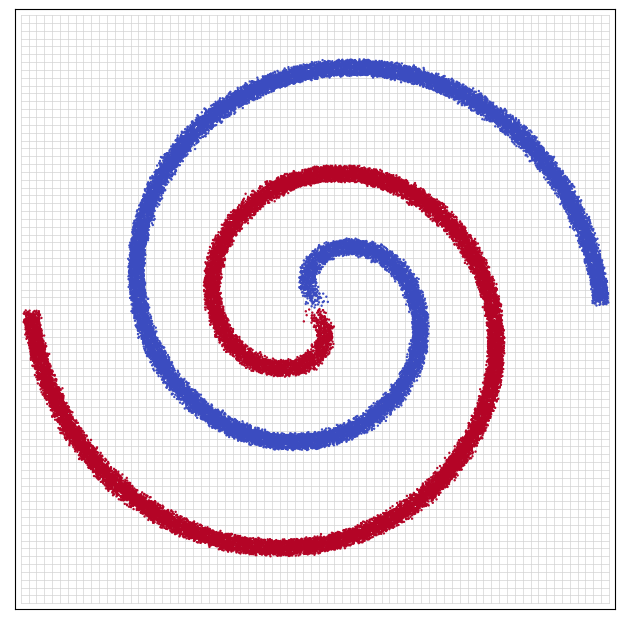}} &
    \includegraphics[width=0.6\textwidth]{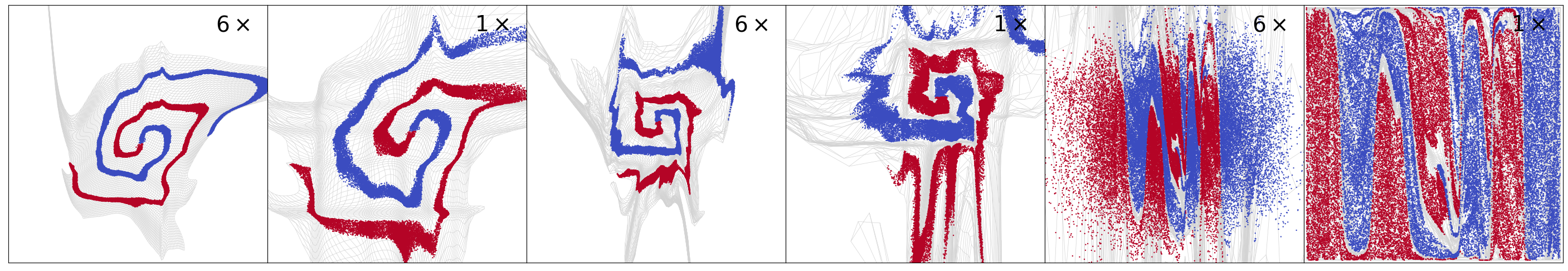} &
    \raisebox{0.02\height}{\includegraphics[width=0.102\textwidth]{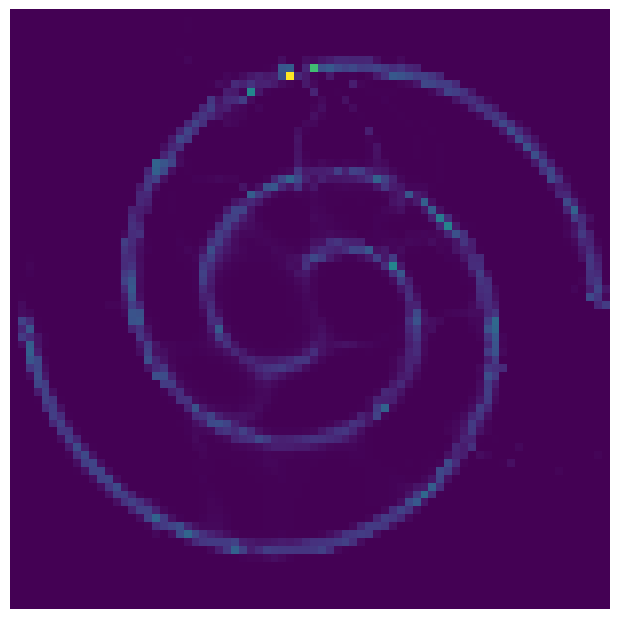}} \\
  \end{tabular}
  \caption{Illustration of diffeomorphic deformations on 2D toy problems. \textbf{Leftmost:} Samples from empirical distribution.
  \textbf{Middle:} Data distribution of intermediate mappings. The symbol "$N\times$" in each grid denotes that the displayed domain
  is $N$ times of the leftmost column. \textbf{Rightmost:} Density estimation by flow. The intermediate mappings reshape the data
  distribution by iteratively performing non-uniform stretching (expansive mapping) and compression (contractive mapping) on its domain.}
  \label{fig:2d_toy}
\end{figure*}
\subsection{Lipschitzness}
As a result of the previous section, $\bbf$ is Lipschitz constrained to bound the gradient, but the constant $K$ is flexible to choose.
This flexibility brings a trade-off between the expressivity of the mapping and the stability of optimization. Because a tight bound is
harmful to the expressivity (by Eq.~\eqref{eq:bound}), while a loose bound may lead to the collapse of
optimization (by Proposition~\ref{proposition:logdet-bound}). One choice of dealing with this trade-off is to explicitly define the
constant $K$ as a hyperparameter when constructing model, e.g., use spectral normalization to control the value of $K$ for linear
maps~\citep{miyato2018spectral}, and fine-tune the $K$ during training. This is difficult because additional effort is needed
and $K$ is not always controllable. Another choice is to limit the variance of the hidden output of model, as its change is approximately
proportional to the change of determinant. Methods in this way include
(1) using contractive activations those have zero derivative at region far from the origin, such as tanh and normal CDF,
(2) adding additional regularization such as $L_2$ transport cost~\citep{onken2020ot}, and
(3) applying a prior distribution to the parameter of model.

\subsection{Blockwise Volume-preserving Initialization}
The initialization of flows has non-negligible impact on the convergence. If every layer was initialized as
a contraction (an expansion) mapping, after the cumulative effect of multiple layers, the variance of hidden layers may become
very small (large), which may slow down or even stop the convergence. A widely used strategy is the volume-preserving initialization,
where every layer is initialized to have zero log-determinant, e.g., initializing the weight of linear mapping with identity or
orthogonal matrix. We can extend this strategy to blockwise volume-preserving initialization, where in each block an expansion
initialization is followed by a contraction initialization, but their combined effect still has zero log-determinant.
The idea of this strategy is from our observation that a map with positive log-determinant is usually followed by a map with negative
log-determinant in a trained flow (Fig.~\ref{fig:loss_curve} (c)).

\subsection{Multimodality}
Let us denote $\sigma(\tau)$ as the number of local maxima of function $\tau:\mathbb{R}^d\to\mathbb{R}$.
We say $\tau$ is multimodal if $\sigma(\tau)>1$ on its domain.
With this definition, we can roughly evaluate the complexity of a probability distribution by $\sigma(p(\bx))$,
and the capacity of a normalizing flow by $\max_{\btheta}\sigma(J_{\bg}(\bz;\btheta))$.
If a target distribution $p(\bx)$ can be perfectly modeled by a flow $\bg_{\btheta}(\bz)$,
it must satisfy $\sigma(p(\bx))\le\max_{\btheta}\sigma(J_{\bg}(\bz;\btheta))$.
The intuition behind this is, the modes of density function that a flow is able to provide over its parameter space should not
be less than the ones of target distribution, or it is impossible for the model to fit the target distribution perfectly.
Therefore, to design a powerful flow, aside from increasing the depth of the model, one approach is to improve the capacity of each
intermediate layer to provide multimodality. For example, using a mixture of logistics~\citep{ho2019flowpp},
or monotonic spline interpolation~\citep{muller2019neural,durkan2019nsf}.

Summarily, normalizing flow can be considered as a sequence of deformations (such as stretching and contraction) between two manifolds,
but the distortion at each step is limited so no tearing and gluing appear (the Lipschitz condition).
See Fig.~\ref{fig:2d_toy} for an illustration. For designing a universal flow, the limit exists by improving the capacity of 
single mapping. But fortunately, the combination of mappings can provide high modality thus breaking the limit.

\section{Proposed Flow}
To exam our theoretical results, we propose a new flow in this section. Our model is built on the work
of RealNVP~\citep{dinh2016density}, Glow and NSF~\citep{durkan2019nsf}. The basic block of our model includes
three components: (\romannumeral1) a linear layer to enhance the interaction of dimensions, (\romannumeral2) two consecutive coupling
layers to capture the information of every dimension, and (\romannumeral3) an elementwise activation with multimodal derivative to
improve the nonlinearity. Each block is parameterized by special structure to satisfy the Lipschitz constraint and initialized with
blockwise volume-preserving strategy. A set of building blocks are then combined together with a multi-scale architecture.

\subsection{Building Block}
\paragraph{Linear layer via invertible convolution.}
We extend the invertible 1$\times$1 convolution~\citep{kingma2018glow} to $k\times k$ convolution with $k$ strides and $k^2$ times
output channels compared to input. Such convolution is an invertible linear mapping if the weight of convolution is invertible,
denoted as $\bbf_\text{conv}:\mathbb{R}^d\to\mathbb{R}^d$.
The map $\bbf_\text{conv}$ can be considered as a fused operation of squeezing and 1$\times$1 convolution if $k>1$. For instance,
for a $h\times w\times c$ tensor, we have a $\frac{h}{k}\times\frac{w}{k}\times k^2c$ output after the transform of $\bbf_\text{conv}$.
The determinant contributed by this convolution is equal to $hw/k^2\det(\bm{w})$ in which $\bm{w}$ is its weight matrix.
In our implementation, the weight is parameterized by a $k^2c\times k^2c$ square matrix and initialized as $\kappa\mathcal{I}$,
where $\mathcal{I}$ is an identity matrix, and $\kappa$ a positive number to be discussed below.

\paragraph{Dual affine coupling layer.}
To let every dimension be transformed in an individual block, we stack two affine coupling layers~\citep{dinh2016density} together.
More precisely, for a partition $\bx = (\bx_1, \bx_2), \bx_1\in \mathbb{R}^r,\bx_2\in\mathbb{R}^{d-r}, 1\le r < d$,
we define a layer of dual affine coupling $\bbf_\text{aff}:\mathbb{R}^d\to \mathbb{R}^d$ by
\begin{equation}
\begin{aligned}
  \by_1 &= \bs_1(\bx_2) \odot \bx_1 + \bt_1(\bx_2), \\
  \by_2 &= \bs_2(\by_1) \odot \bx_2 + \bt_2(\by_1),
\end{aligned}
\end{equation}
where $\bs_1,\bt_1:\mathbb{R}^{d-r}\to\mathbb{R}^r$ and $\bs_2,\bt_2:\mathbb{R}^r\to\mathbb{R}^{d-r}$ are maps parameterized
by neural networks, $\odot$ is the Hadamard product. In particular, it follows $\bs > 0$ to ensure the invertibility of $\bbf_\text{aff}$.
The determinant $\det(J_{\bbf_\text{aff}})$ is simply the product of every entries of $\bs_1$ and $\bs_2$.
To control the initialization and the output range of $\bs$, the last layer of neural networks is a linear layer followed by a 
mixture of $M$ Mexican-hat-like activations:
\begin{equation}
\begin{aligned}
  \varphi_m(\bx) &= w_m\phi(\bx) + b_m, \\
  \log\bs &= \frac{1}{M}\sum_{m=1}^M(1-\varphi_m^2)\exp(-\frac{\varphi_m^2}{2}),
  \label{eq:mexican}
\end{aligned}
\end{equation}
where $\phi$ is the pre-output of neural network, $w_m$ the weight matrix initialized to 0, and $b_m$ the bias term whose initialization
follows the blockwise volume-preserving strategy. With this layer, $\bs$ is limited in the range $(0.5, 3)$.

\paragraph{Monotonic rational-quadratic activation.}
To improve the multimodality of determinant mapping $\tau$, we construct a monotonic and elementwise activation function
$f_\text{act}:\mathbb{R}\to\mathbb{R}$ using rational-quadratic splines~\citep{gregory1982piecewise,durkan2019nsf}.
The function $f_\text{act}$ is defined in the interval from $[x^{(0)}, x^{(I)}]$ to $[y^{(0)}, y^{(I)}]$,
and is fully described by $3(I+1)$ parameters $\{(x^{(i)}, y^{(i)}, \alpha^{(i)})\}_{i=0}^I$ which
satisfy $x^{(i)}<x^{(i+1)}, y^{(i)}<y^{(i+1)}$ and positive derivatives at the boundaries $\alpha^{(i)}>0$.
In each sub-interval (a.k.a. a bin) $[x^{(i)},x^{(i+1)}]$, $f_\text{act}$ is obtained by a rational-quadratic function.
Let bin width $\Delta_x^{(i)}=x^{(i+1)}-x^{(i)}$, bin height $\Delta_y^{(i)}=y^{(i+1)}-y^{(i)}$,
and bin ratio $\delta^{(i)}=\Delta_y^{(i)}/\Delta_x^{(i)}$, for a point $x\in [x^{(i)}, x^{(i+1)}]$,
denote $\xi(x;i)=(x-x^{(i)})/\Delta_x^{(i)}$ in which $0\le\xi\le 1$, we have
\begin{equation}
  f_\text{act}(\xi;i)=y^{(i)}+\frac{\Delta_y^{(i)}\left[\delta^{(i)}\xi^2+\alpha^{(i)}\xi(1-\xi)\right]}{\delta^{(i)}+\rho^{(i)}\xi(1-\xi)},
\end{equation}
where $\rho^{(i)}=\alpha^{(i+1)} + \alpha^{(i)} - 2\delta^{(i)}$, for $i=0,1,\cdots,I-1$. The function $f_\text{act}$
is invertible with analytical inverse, and its first derivative is computationally tractable as
\begin{equation}
  \frac{d}{dx}f_\text{act} = \frac{\gamma^{(i)}(\delta^{(i)})^2}{[\delta^{(i)}+\rho^{(i)}\xi(1-\xi)]^2},
\end{equation}
where $\gamma^{(i)}=\alpha^{(i+1)}\xi^2 + 2\delta^{(i)}\xi(1-\xi) + \alpha^{(i)}(1-\xi)^2$.
This derivative can be multimodal as $\max\sigma(df_\text{act}/dx)=I$, when it can also be trivial as a constant
if $\alpha^{(i)}=\delta^{(i)}=\beta$ is a same value for all $i$. In the case of the latter, $\rho^{(i)}=0$ and $f_\text{act}$
degenerates to a linear function with slope $df_\text{act}/dx=\gamma^{(i)}=\beta$.

The above three components are composed together as the building block of our model as
$\bbf_l = \bbf_\text{act}\circ\bbf_\text{aff}\circ\bbf_\text{conv}$.
To follow the principles discussed in Sec.~\ref{sec:principles}, we wish the first two components $\bbf_\text{aff}\circ\bbf_\text{conv}$
to act as an expansive function with loose constraint. Here the constraint is flexible but important for training stability,
and it is one of the reasons we limit the range of $\bs$ by Eq.~\eqref{eq:mexican}.
For $f_\text{act}$, we let it be contractive by limiting $\alpha^{(i)},\delta^{(i)}\in (0,1)$, which conditions are sufficient to let
$df_\text{act}/dx\le 1$ holds almost everywhere in its domain. To follow the strategy of blockwise volume-preserving initialization,
we initialize every components as linear maps satisfying $\kappa(\bs)_j=1/\beta$ by controlling the initialization of parameters
$\kappa,b_m$ and $(x^{(i)},y^{(i)}, \alpha^{(i)})$. Thus $\bbf_l$ is initialized as a blockwise identity function with determinant:
\begin{equation}
  \det(J_{\bbf_\text{conv}})\det(J_{\bbf_\text{aff}}) =\frac{1}{\det(J_{\bbf_\text{act}})}=\beta^{-d},
\end{equation}
in which $\beta$ is a hyperparameter in $(0,1]$, experically a narrower range $[0.5, 1]$ is advised.

\subsection{Multi-scale Architecture}
The multi-scale architecture introduced by~\citet{dinh2016density} is a framework for the composition of multiple diffeomorphic transforms,
where it factors out half of the dimensions at each scale. Formally, multi-scale architecture with two levels is a composition of
diffeomorphisms as follows.
\begin{equation}
\begin{aligned}
  \bbf = \bbf_2\circ\bbf_1:\mathbb{R}^d &\to\mathbb{R}^d, \text{ where } \bbf_2=(\text{id}, \bbf_3), \\
  \text{id}: \bx\in\mathbb{R}^r &\to\bx,\  \bbf_3: \mathbb{R}^{d-r}\to\mathbb{R}^{d-r}.
\end{aligned}
\end{equation}
There $r$ is specified to $r=d/2$ in~\citet{dinh2016density}, while it can vary in range $[1,d)$. The function $\bbf_3$
could be another composition of diffeomorphisms in the same way, resulting in an architecture with multiple levels.
An interesting property of multi-scale architecture is the importance ranking effect between dimensions, i.e.,
compared to the latent features generated by the identity mapping in $\bbf_2$, the ones generated by the non-identity mapping $\bbf_3$
carry a different amount of information for reconstructing the input. This is a property shared with PPCA,
known as dimensionality reduction. See our experiment in Sec.~\ref{sec:multiscale}.

\section{Related work}
Related topics to our work include:
(1) Tractability, in which the computational challenge of Jacobian determinant is concerned.
(2) Expressivity, in which the construction of diffeomorphic mapping with restricted determinant is concerned. 
For these two topics, we refer to the surveys by~\citet{papamakarios2019normalizing} and~\citet{kobyzev2020normalizing}.

\paragraph{Universality.} As a complement of our work, \citet{teshima2020coupling} proves affine coupling flows (a case of QLF) are
universal approximators in the sense of $L^p$ norms, and~\citet{koehler2020rep} further shows the relation of universality with the
depth of affine couplings. The universality for other invertible architectures is also explored
by~\citet{zhang2020approximation} (on continuous flows) and~\citet{kong2020expressive} (on matrix determinant lemma-based flows).

\paragraph{Flows via Optimal Transport.} Optimal Transport (OT) provides a different measure of statistical distance from MLE,
which encourages straight trajectories between two distributions~\citep{onken2020ot} and in some sense penalizes unreasonable increase
of volume. Recent work of~\citet{zhang2018monge}, \citet{finlay2020train}, \citet{yang2020potential} and~\citet{onken2020ot}
introduced OT into normalizing flows and found the training stability is improved. It would be interesting to further investigate
whether QLF or other restricted flows have closed form formulas under the OT metric in future work.

\section{Experiments}

\begin{table}[t]
\caption{Density estimations of standard benchmarks in bits/dim (lower is better).
The number in brackets is the number of parameters ($\times 10^6$). The results for QLF$^\ast$ are theoretical
value calculated by equation~\eqref{eq:maxll}.}
\vskip 0.15in
\begin{center}
\begin{small}
\begin{sc}
\begin{tabular}{lcccr}
  \toprule
  model & cifar10 & imagenet 32 & celeba 1024 \\
  \midrule
  realnvp  & 3.49 & 4.28 & - \\
  glow      & 3.35 (44.0) & 4.09 & - \\
  rq-nsf    & 3.38 (11.8) & - & - \\
  \midrule
  qlf$^\ast$       & 2.01 & 1.75 &  - \\
  ours      & 3.37 (12.6) & 4.03 & 0.64 (23.3) \\
  \bottomrule
\end{tabular}
\end{sc}
\end{small}
\end{center}
\vskip -0.1in
\label{tab:density}
\end{table}

\subsection{Density Estimation}
To evaluate our proposed flow in density estimation, we train it on standard image benchmarks
CIFAR10~\citep{krizhevsky2009learning}, downsampled 32$\times$32 ImageNet~\citep{russakovsky2015imagenet}
and CelebA-HQ 1024$\times$1024~\citep{karras2017progressive}.
The results in Table.~\ref{tab:density} demonstrate that the performance of our model is similar to RQ-NSF on CIFAR10,
and slightly better than Glow on ImageNet 32$\times$32. Benefit from the multimodal activation, compared to Glow, significantly
less parameters is required to achieve the same score. In addition,
our model is scalable to CelebA-HQ 1024$\times$1024, while the others are absent in this dataset due to numerical issues or memory limitation.
To the best of our knowledge, it is the first flow-base model in the literature that can train on this scale.
In particular, the values of QLF show the gap between theoretical and experimental optima, providing a useful guidance for model design.

\begin{figure*}[t]
  \centering
  \subfigure[NLL curve]{\includegraphics[width=0.31\linewidth]{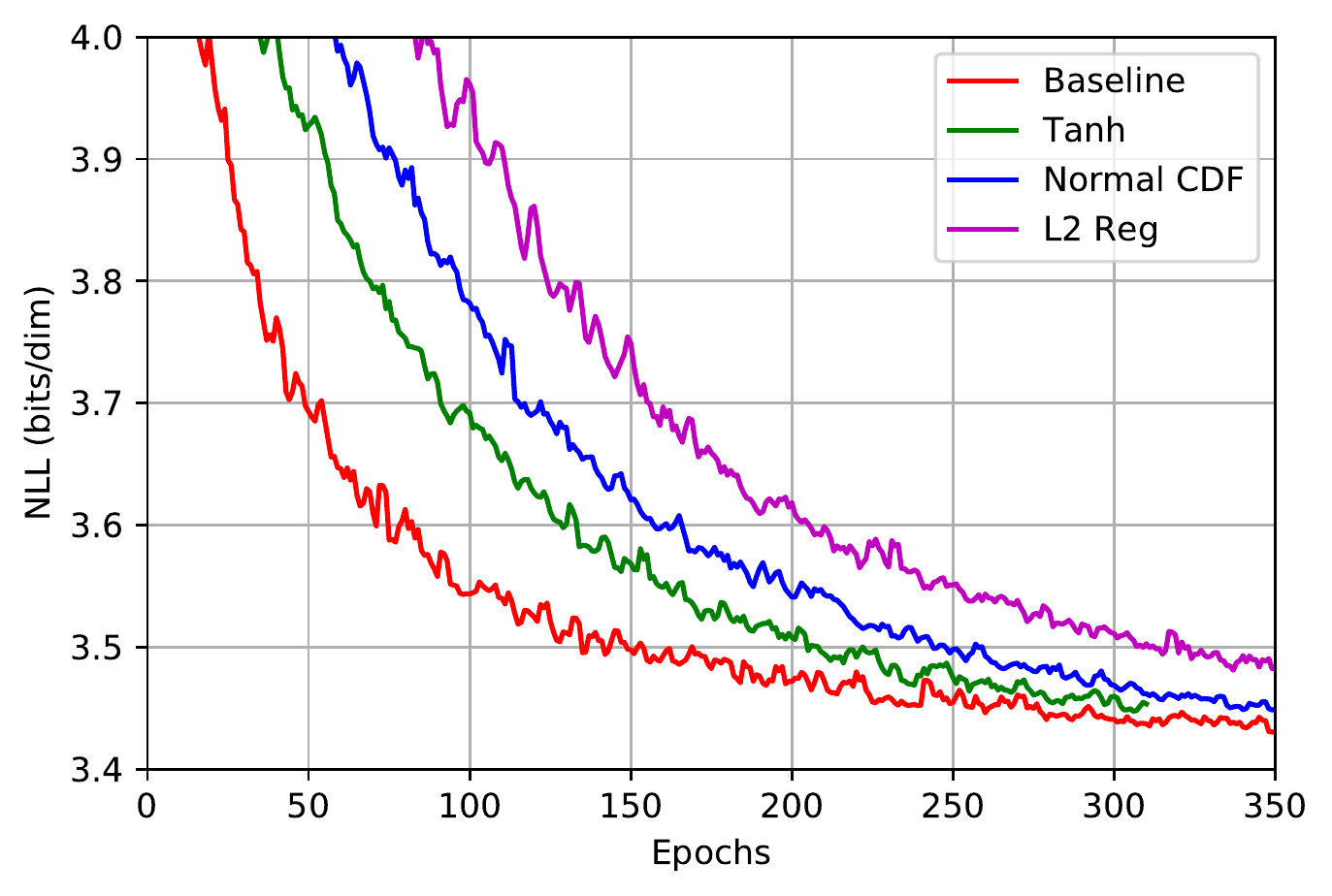}}
  \subfigure[Variance distribution]{\includegraphics[width=0.31\linewidth]{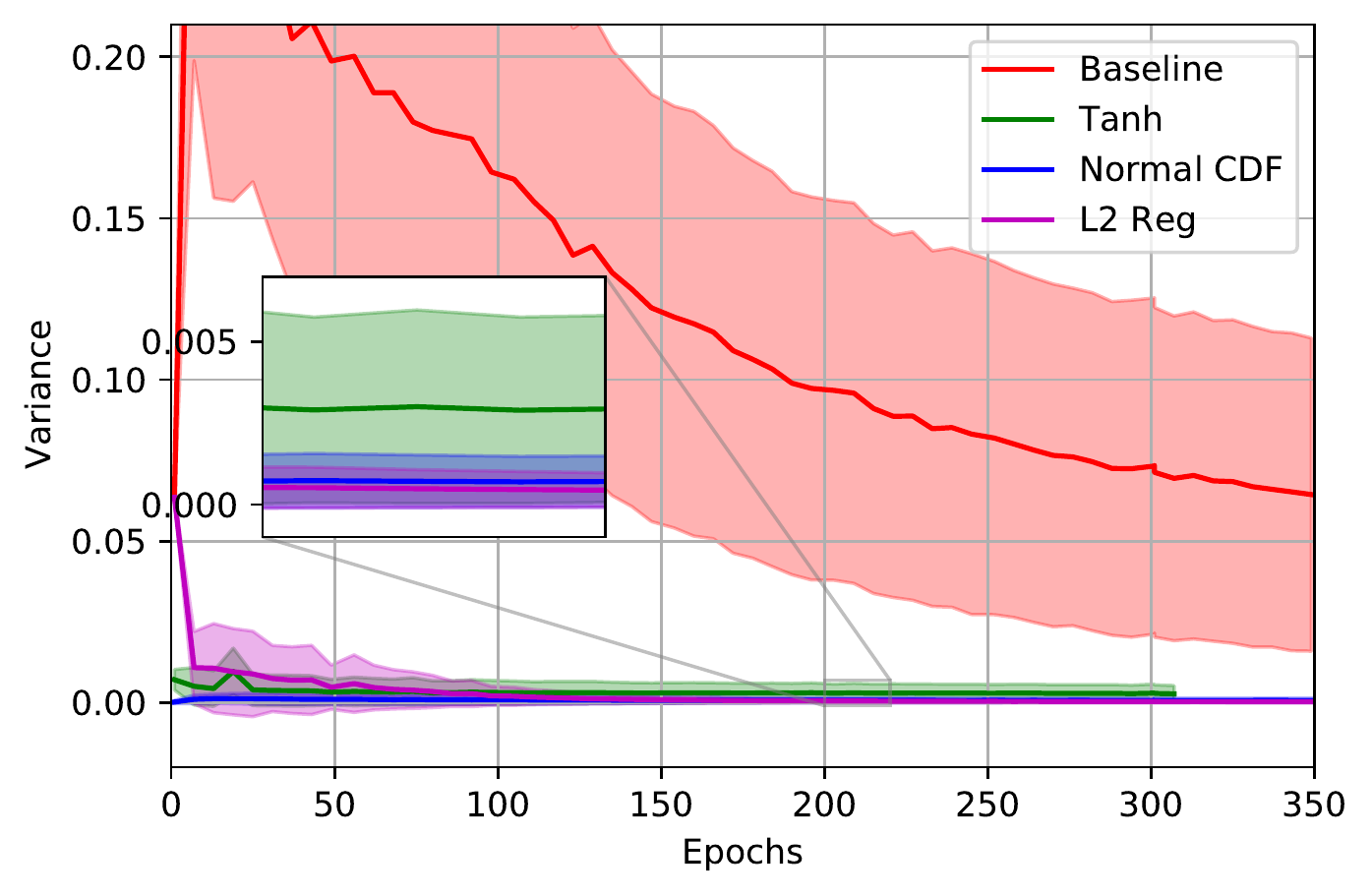}}
  \subfigure[Log-determinant]{\includegraphics[width=0.31\linewidth]{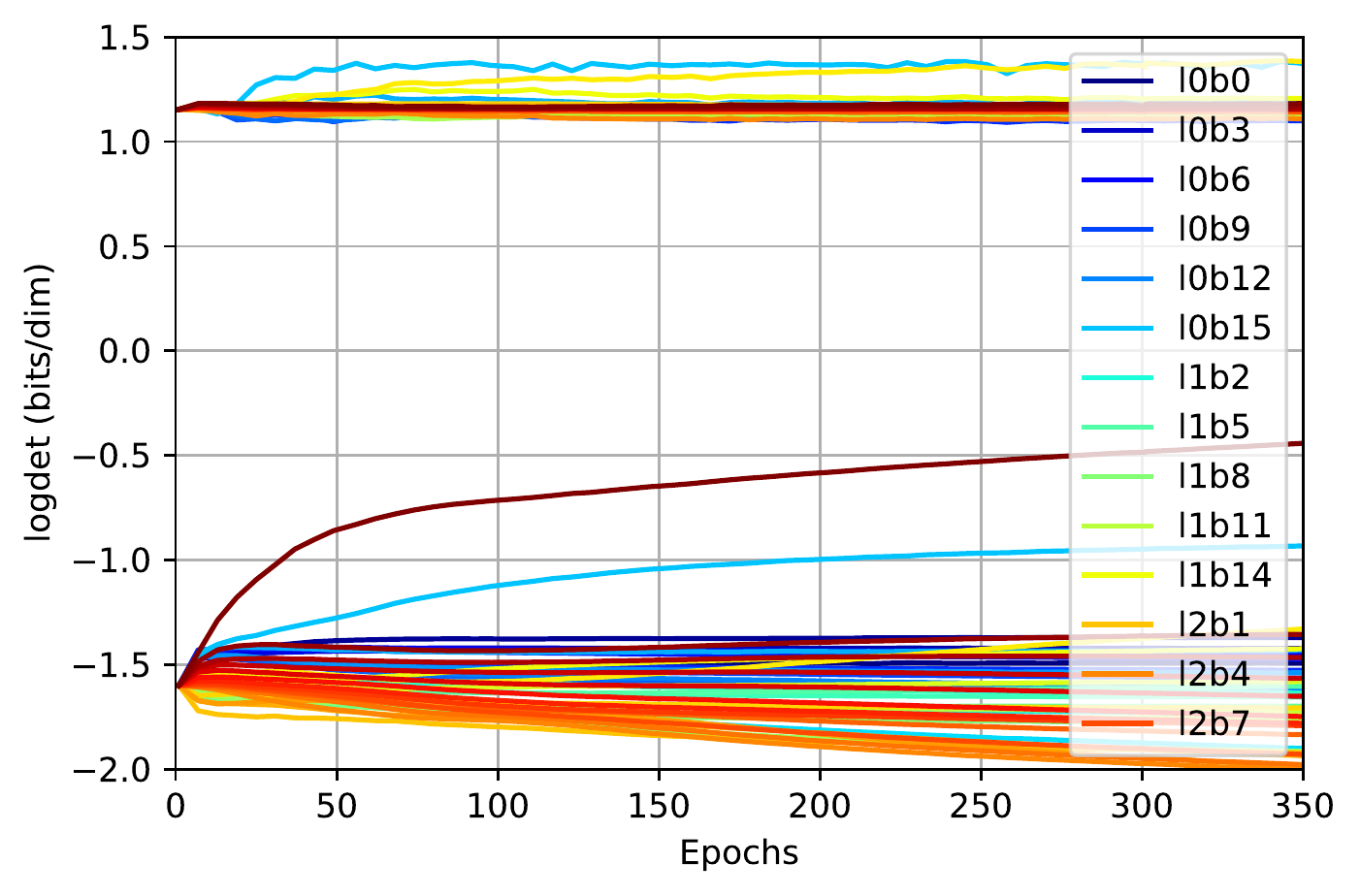}}
  \caption{Training curves on CIFAR10. (a) Negative log-likelihood (NLL) curves.
  (b) The distribution of the variance for all intermediate mappings.
  (c) The log-determinant of each intermediate mapping, an expassive layer followed by a contractive layer.}
  \label{fig:loss_curve}
\end{figure*}

\subsection{Effect of Lipschitzness}
We explore the effect of Lipschitzness on CIFAR10. 
In Fig.~\ref{fig:loss_curve}, the baseline is our proposed flow. As a comparison, contractive activations Tanh and normal CDF are inserted
between $\bbf_\text{aff}$ and $\bbf_\text{act}$, and $L_2$ transport cost is applied to the output of $\bbf_\text{act}$, respectively.
These added activations or regularity have tighter constraint than the baseline. And they lead to slower convergence and
worse NLL estimation (Fig.~\ref{fig:loss_curve} (a)). For a model having worse NLL performance, it also has smaller
averaged variance over all intermediate maps (Fig.~\ref{fig:loss_curve} (b)). In contrast, the model without any constraint on
$\bbf_\text{aff}$ and $\bbf_\text{act}$ converges faster and has higher variance at the beginning of training,
but soon collapses due to the gradient problems, thus no training curve is displayed.
This result validates the relationship between Lipschitzness of $\bbf$, boundedness of $\tau$, and variance change by $\bbf$ discussed
in Sec.~\ref{sec:properties}.

\begin{figure}[t]
  \centering
  \setlength{\tabcolsep}{0.8pt}
  \renewcommand\arraystretch{0.6}
  \begin{tabular}{cc}
    \raisebox{0.5\height}{\includegraphics[width=0.1\linewidth]{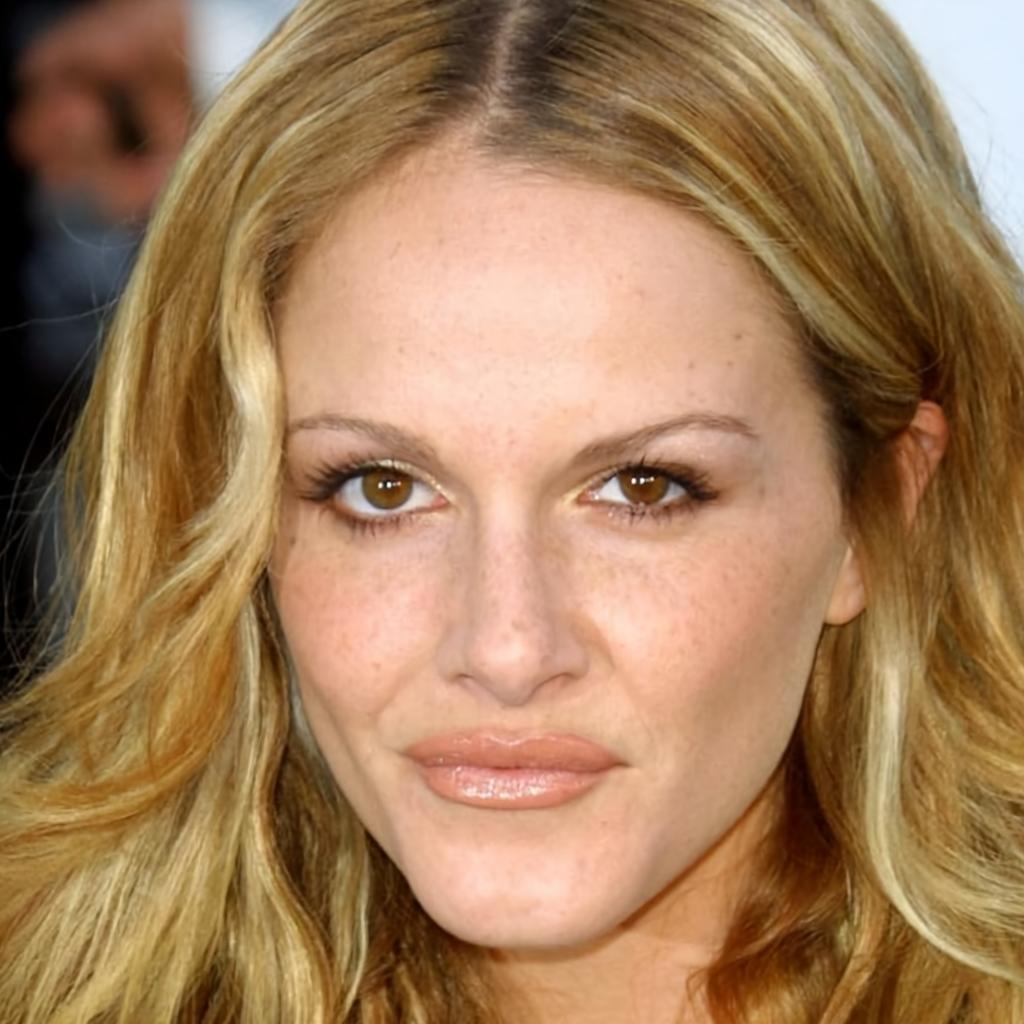}} &
    \includegraphics[width=0.75\linewidth]{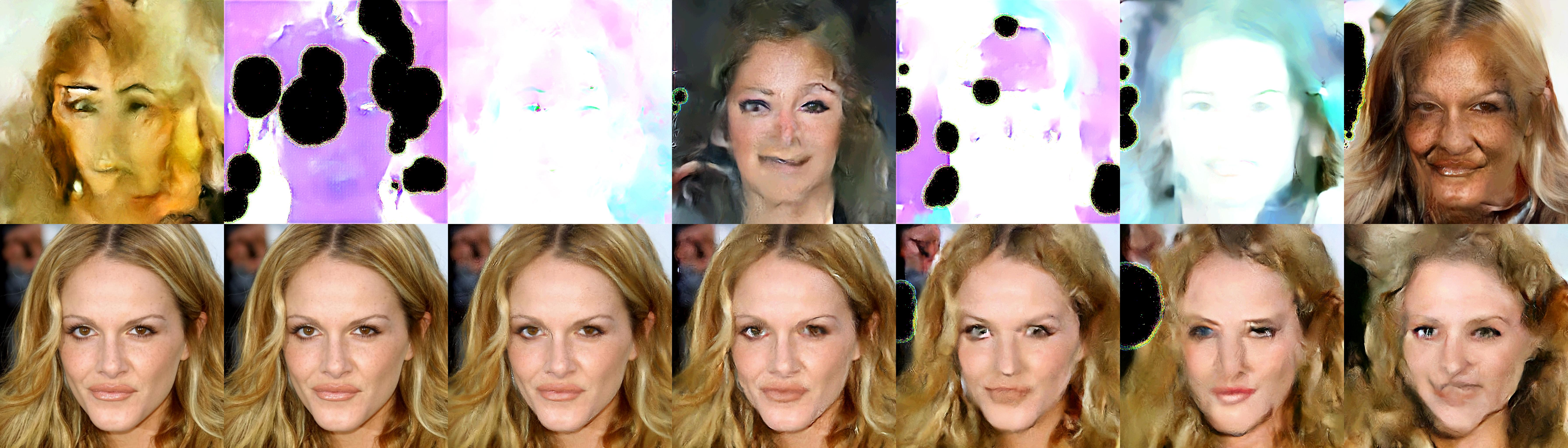} \\
    \raisebox{0.5\height}{\includegraphics[width=0.1\linewidth]{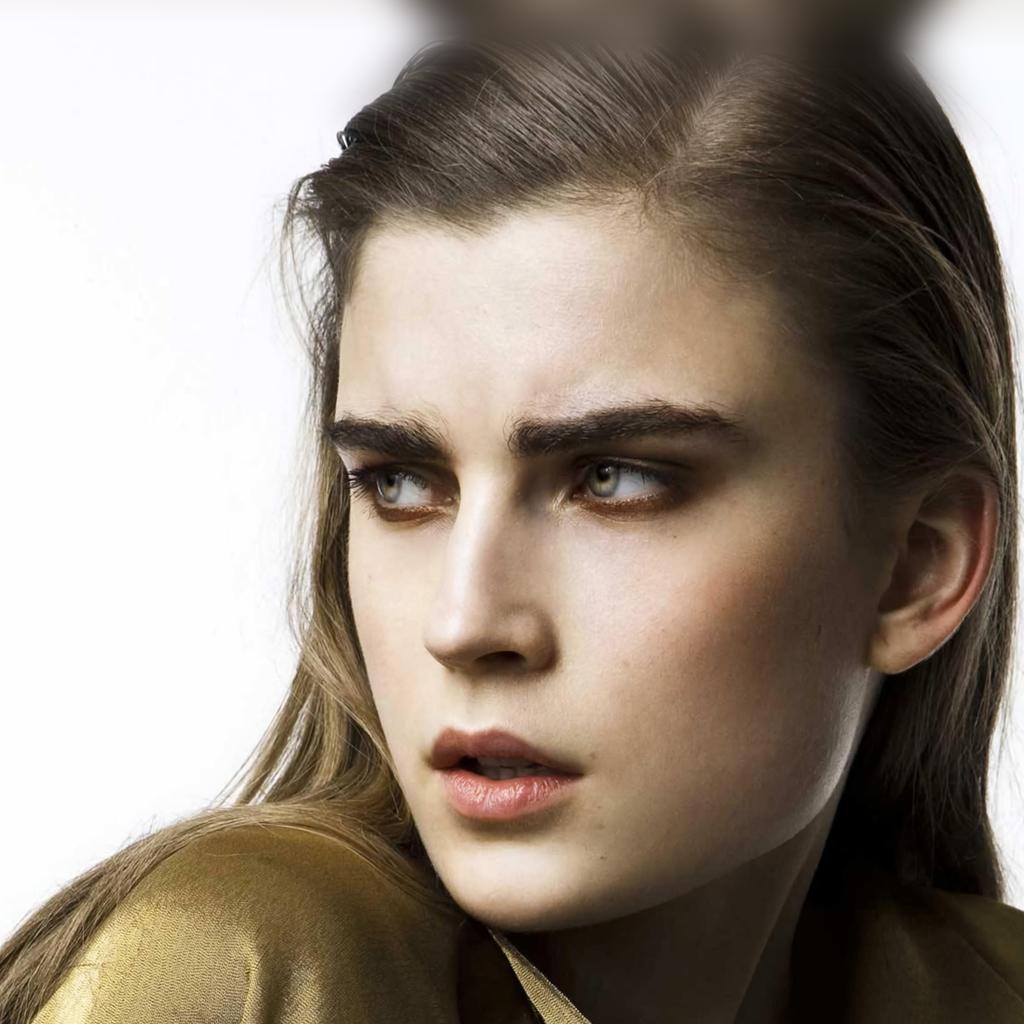}} &
    \includegraphics[width=0.75\linewidth]{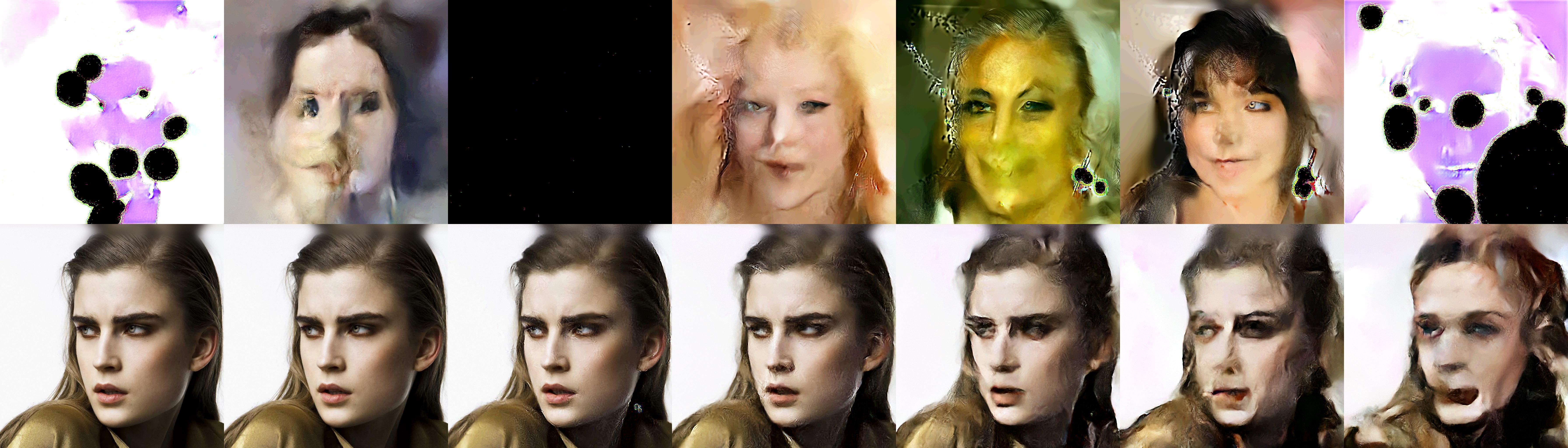} \\
    \raisebox{0.5\height}{\includegraphics[width=0.1\linewidth]{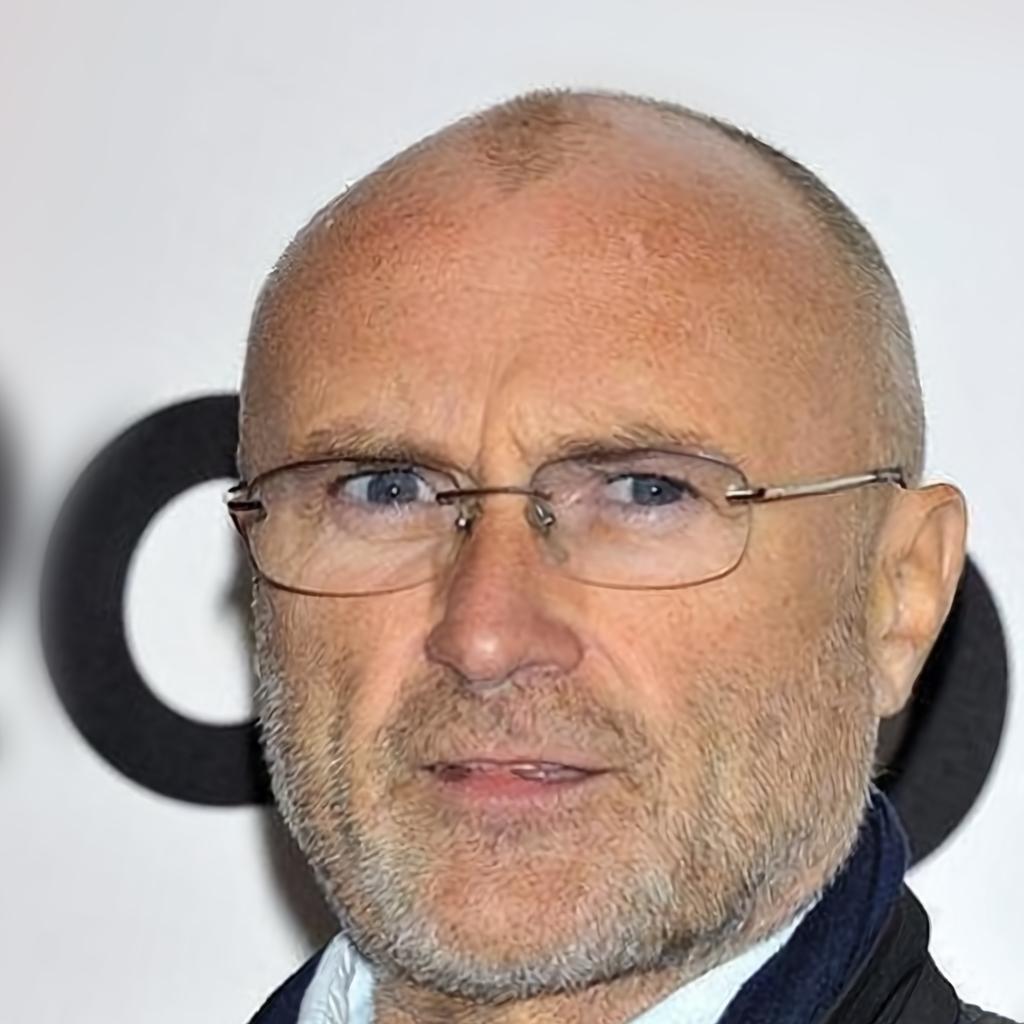}} &
    \includegraphics[width=0.75\linewidth]{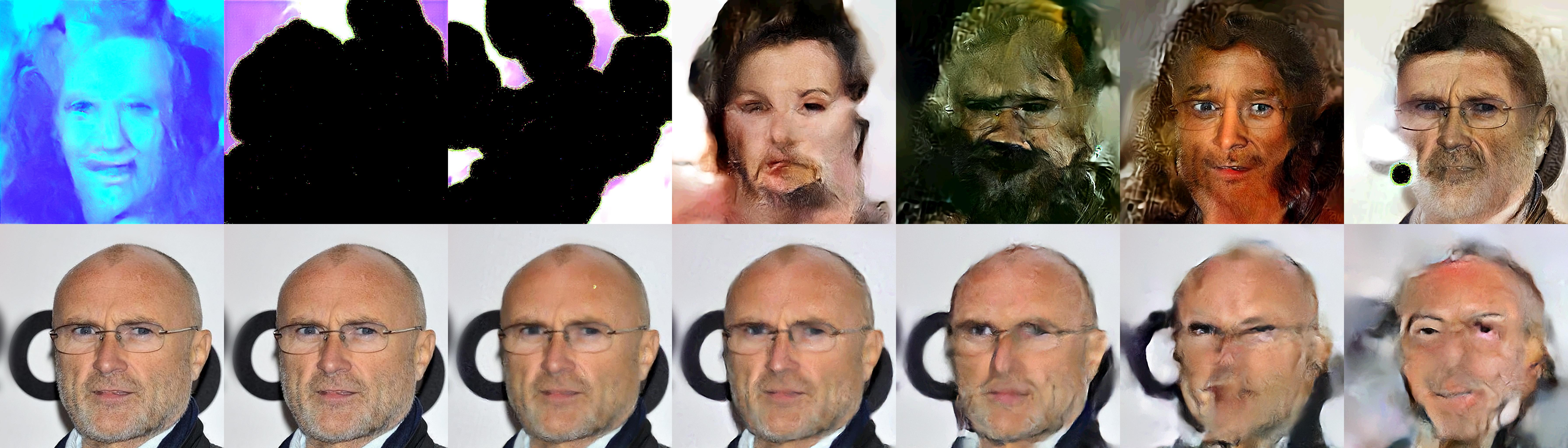}
  \end{tabular}
  \caption{Perturbations of latent variables. The leftmost column represents the original image, while the others are reconstructions.
  More specifically, the reconstruction in the $k$-th column ($k\ge 1$) and the first row relative to each real image is obtained by
  storing the latent variables in the first $k$ levels and resampling the others, while the second row
  resampling the first $k$ levels and storing the others. Latent variables in an earlier layer have smaller level order.
  The results indicate the latent variables in higher levels contain more information about the data.}
  \label{fig:perturb}
\end{figure}

\subsection{Multi-scale Architecture for Dimensionality Reduction}
\label{sec:multiscale}
In Fig.~\ref{fig:perturb}, we use an 8 levels model pretrained on CelebA-HQ 1024$\times$1024 to obtain the latent variables
of real image, then partially resample these variables from a noise distribution, and finally decode the perturbed variables to get
the reconstruction image. The results show that the latent variables in a higher level are more informative for the reconstruction.
This can be explained as follows. Group the latent variables $\bz=(\bz_1,\bz_2,\cdots)$ by the corresponding level order,
and consider the determinant map on generation direction $\pi(\bz)=\tau(\bg(\bz)):\bz\mapsto|\det_{J_{\bg}}(\bz)|$ as a hypersurface
embedded in $\mathbb{R}^{d+1}$, when varying the dimensions in the $i$-th level but fixing the others, a new hypersurface
$\pi_i:\by\mapsto\pi(\bz_{i-1},\by,\bz_{i+1})$ in the subspace is generated. Due to the careful design of multi-scale architecture,
the function $\pi_i$ with lower level $i$ has less layer thus is less complex. This results in that the hypersurface $\pi$ along the
directions of the $i$-th group of dimensions are smoother, which dimensions containing less information about the dataset.
This property is useful for dimensionality reduction and data compression, specifically for the scenario of data transmission
by extracting latent variables in high levels and reconstructing at the remote.

\section{Conclusion}
This work demonstrates the uniqueness of Jacobian determinant mapping through Radon-Nikodym theorem,
and shows the closed form exists for flows in the form of QLF. For the training of flows, the convergence condition is given.
In addition, a new flow is proposed and its improved stability and scalability are validated by numerical experiments.
Normalizing flow is essentially a nonlinear extension of PPCA, so it has the potential for applications where PPCA is
applicable, such as dimensionality reduction and data reconstruction.


\bibliography{refs}
\bibliographystyle{icml2021}

\appendix
\onecolumn
\newpage
\section{Quasi-Linear Flow}
\subsection{The Stationary Point of QLF}
Suppose the determinant of $\bW$ is positive. To obtain the stationary point of QLF, we can calculate the gradient of
Eq.~\eqref{eq:loglikelihood} with respect to $\bW$ at point $\bx$:
\begin{equation}
  \fpartial{\mathcal{L}}{\bW}=-\bW\bS^T+(\bW^{-1})^T
\end{equation}
Note that $\bW$ is invertible. At the stationary point, $\bS=\bW^{-1}(\bW^{-1})^T$. Since $\bS$ is symmetric, applying eigendecomposition, we
have $\bS=\bb{V}\Lambda\bb{V}^T$, in which $\bb{V}$ is orthogonal whose columns are the eigenvectors of $\bS$ and $\Lambda$ is diagonal with
the corresponding eigenvalues. Therefore, there is a solution when $\bW=\bb{U}\Lambda^{-1/2}\bb{V}^T$ with $\bb{U}$ being an arbitrary
orthogonal matrix.

\subsection{Comparison with PPCA}
For the case of $\bW$ and $\bb{b}$ being independent of $\bx$, 
Eq.~\eqref{eq:loglikelihood} can be rewritten as
\begin{equation}
 \mathcal{L} = -\frac{1}{2}\left\{d\log 2\pi\!+\!\tr(\bM\bS^\prime)\!+\!\log|\det(\bM^{-1})|\right\},
 \label{eq:ppca_ll}
\end{equation}
in which $\bS^\prime=\mathbb{E}_{p_{\bm{X}}}[\bx\bx^T]$ is the sample covariance matrix of the observations (supposing $\bx$ is zero-mean).
In this case, it can be shown that Eq.~\eqref{eq:ppca_ll} is maximized when $\bW=\bb{U}\Lambda^{-1/2}\bb{V}^T$, where $\bb{V}$ and $\Lambda$
are the eigenvectors and eigenvalues of $\bS^\prime$ respectively. Restricting all the smallest $d-r$ eigenvalues to be $\sigma^2$ and
separating them from $\Lambda$, then the corresponding latent variables form the noise term $\epsilon\sim \mathbb{N}(0, \sigma^2 I)$ in the
original PPCA~\citep{Tipping99probabilisticpca}, while the other $r$ terms are the principal components.

\section{Proofs}
\begin{proof} \textbf{Inequality~\eqref{eq:bound}}

The function $\bbf$ is assumed to be differentiable and $K$-Lipschitz continuous. Therefore, by difinition, for every normalized
eigenvector $\be_i$ and $\varepsilon>0$, there exists $\delta>0$ such that if $h <\delta$ we have
\begin{equation}
  \left|\frac{\|\bbf(\bx+h\be_i)-\bbf(\bx)\|}{h}-\|J_{\bbf}(\bx)\be_i\|\right| < \varepsilon.
\end{equation}
Therefore
\begin{equation}
  \|J_{\bbf}(\bx)\be_i\| < \frac{\|\bbf(\bx+h\be_i)-\bbf(\bx)\|}{h} + \varepsilon < K +\varepsilon.
\end{equation}
Since this holds for every $\varepsilon$, we have $\|J_{\bbf}(\bx)\be_i\|<K$. By induced matrix norm $\|J_{\bbf}(\bx)\|:=\sup_{\|\be_i\|=1}\|J_{\bbf}(\bx)\be_i\|$, we have
\begin{equation}
  \|J_{\bbf}(\bx)\be_i\|\le \|J_{\bbf}(\bx)\|\le K,
\end{equation}
Furthermore, by Hadamard's inequality, $|\det(J_{\bbf}(\bx))|\leq \prod_{i=1}^d \|J_{\bbf}(\bx)\be_i\|$. Hence the inequality~\eqref{eq:bound} gets proven.
\end{proof}

\begin{proof}\textbf{Proposition~\ref{proposition:logdet-bound}}

For $l\in\{1,2,\cdots,L\}$, assume $\|\btheta_l\|$ is bounded from above, i.e.,
there exists finite $C>0$ such that $\|\btheta_l\|\le C$ for all $\bx\in\bm{X}$. By Eq.~\eqref{eq:grad_theta}, we have
\begin{equation}
  \overbrace{\|\fpartial{\bh_{l+1}}{\btheta_l}\|\cdot\prod_{\substack{j=l+1\\j\neq i}}^{L}\|J_{\bbf_j}\|}^{\alpha}\cdot\|J_{\bbf_i}\|+o\leq C,\text{ for } l+1\leq i \leq L.
\end{equation}
Note $o$ is non-negative, therefore
\begin{equation}
  \alpha\|J_{\bbf_i}\|\leq C.
  \label{eq:bounded_neq}
\end{equation}
Now consider the term $\alpha$, since $\bbf_l$ is invertible for all $l$,
we have $0<|\det(J_{\bbf_j})|^{1/d}\leq\|J_{\bbf_j}\|$. Moreover, it is reasonable to assume there are finite points such that $\|\fpartial{\bh_{l+1}}{\btheta_l}\|=0$,
so $\alpha>0$ holds almost everywhere. For the case $\alpha>0$, if $\|J_{\bbf_i}\|$ is unbounded, i.e., there exists a point $\bx$ and a real
number $K_i>0$ such that $\alpha\|J_{\bbf_i}(\bx)\|>\alpha K_i$. It holds for every $K_i$, hence we can always find a $K_i$ such that
$\alpha\|J_{\bbf_i}(\bx)\|>C$. However, this contradicts Eq.~\eqref{eq:bounded_neq}, thus $\|J_{\bbf_i}\|$ must
be bounded (by finite $K_i>0$). Easily, $\sum_{l}\log|\det(J_{\bbf_l})|\leq \sum_{l}d\log K_l=d\log\prod_{l}K_l$ is also true.
\end{proof}

\section{Experimental Details}
\paragraph{Neural Networks.} For the implementation of neural networks in dual affine coupling layers, we use a residual
network (ResNet;\citealt{he2016deep}) for CIFAR10 and Imagenet 32$\times$32, and a simple convolutional network (ConvNet) for
CelebA-HQ 1024$\times$1024. Concretely, in ResNet, a 3$\times$3 convolution layer is followed by one residual bottleneck block
with 3 convolution layers (kernel size of 1$\times$1,3$\times$3, 1$\times$1 respectively) gated by channel-wise
attention~\citep{hu2018squeeze}, followed by two 1$\times$1 convolution layers
between which another channel-wise attention is inserted. In ConvNet, due to computational resource constrains, three convolutional layers
with kernel size of 3$\times$3, 1$\times$1, 3$\times$3 respectively are used. The number of hidden units is 128 in ResNet,
and 256 in ConvNet.

\begin{figure}[t]
  \centering
  \setlength{\tabcolsep}{5pt}
  \subfigure[]{\includegraphics[width=0.41\textwidth]{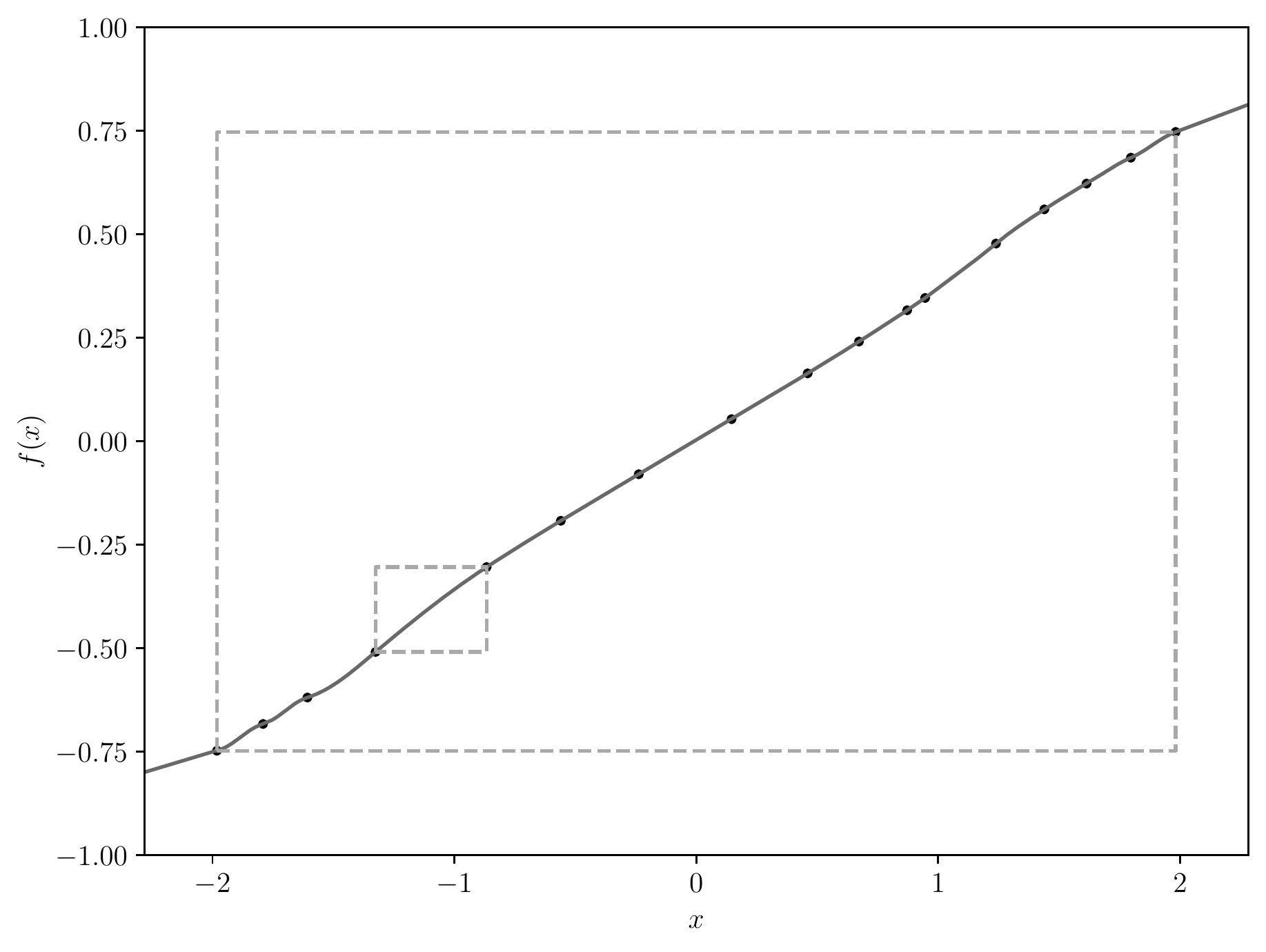}}
  \subfigure[]{\includegraphics[width=0.4\textwidth]{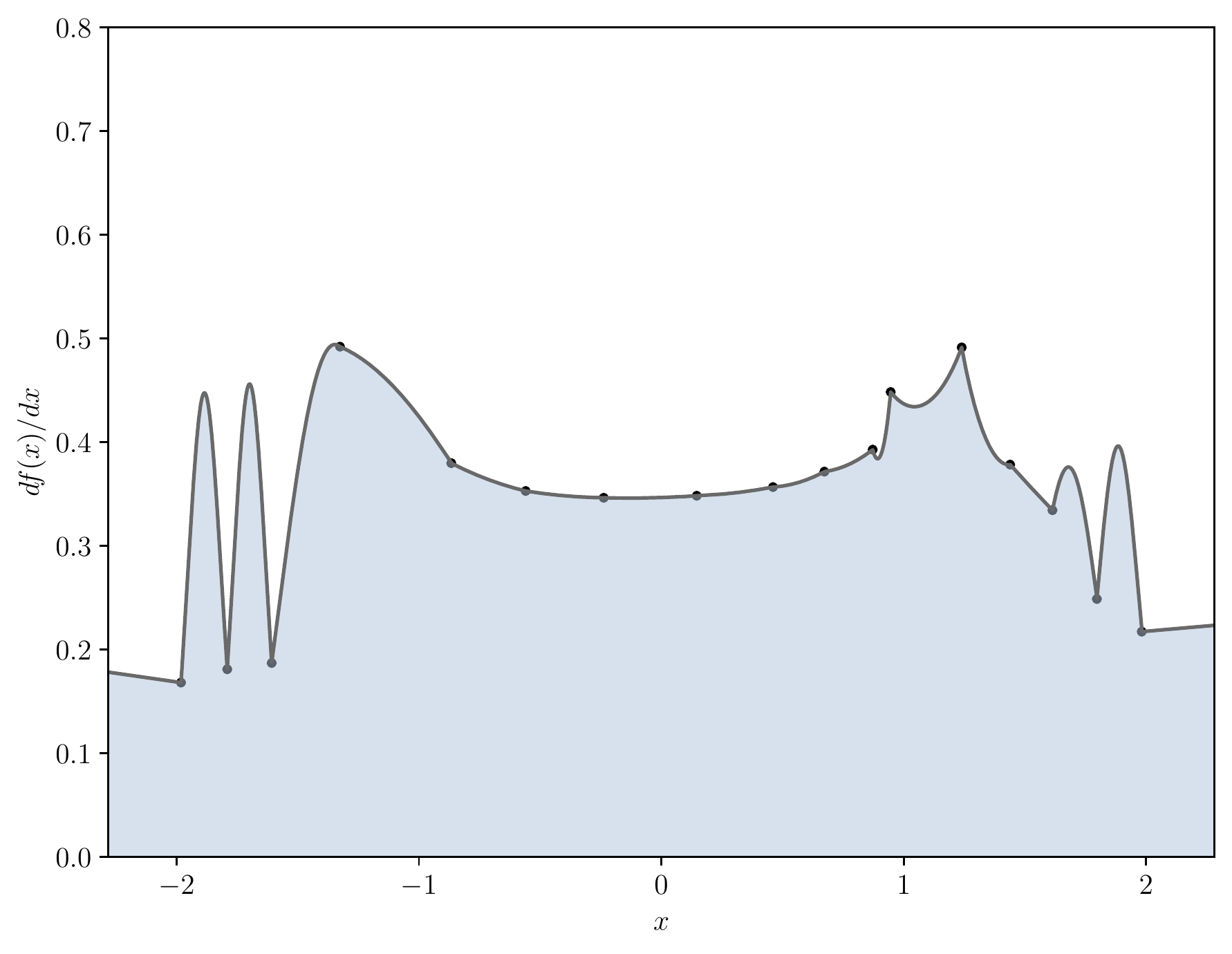}}
  \caption{An example of monotonic rational-quadratic activation (a) and its derivative curve (b).}
  \label{fig:act}
\end{figure}

\paragraph{Rational-quadratic Activation.}
In order to restrict the range of derivative, we have a different implementation from~\citet{durkan2019nsf}. In our implementation,
the knots $\{x^{(i)},y^{(i)}, \alpha^{(i)}\}_{i=0}^{I}$ is parameterized by vectors $\theta_x,\theta_y,\theta_a\in\mathbb{R}^{I+1}$
respectively. The ratio of height and width $\delta^{(i)}$ in each bin (the inner box in
Fig.~\ref{fig:act} (a)) is strictly limited in range $[0,1]$ by
\begin{align}
   \bar{b}_x &= \text{softmax}(\theta_x), \\
   \bar{b}_y &= \text{sigmoid}(\theta_y)\cdot \bar{b}_x, \\
   x^{(i)} &=  (2\times\text{cumsum}(\bar{b}_x)_i - 1)w, \\
   b_y &= \text{cumsum}(\bar{b}_y), \\
   y^{(i)} &=  (2\times(b_y)_i - \max{(b_y)})w,
\end{align}
where cumsum is the cumulative sum of its inputs, and $w$ is a learnable variable that controls the width and position of the outer box
in Fig.~\ref{fig:act} (a). The derivative at each knots is also limited in range $[0,1]$ by
\begin{equation}
\alpha^{(i)}=\text{sigmoid}(\theta_a)_i.
\end{equation}
By this way, the derivative of activation is less than one almost everywhere (Fig.~\ref{fig:act} (b)).
For the region out of the outer box, it is also treated as a bin. For instance, the region $(w,\infty)$ is a bin whose right boundary
is represented by a relative large number (e.g. $1\times 10^5$) and the parameters of right boundary are fixed during training.
In our experiments, the number of bins is $I=16$.

\paragraph{Multi-scale Architecture.} For datasets CIFAR10 and ImageNet 32$\times$32, the model has 3 levels with 16 blocks per level,
and half of the dimensions is splitted out at each level. For CelebA-HQ 1024$\times$1024, the model has 8 levels with 12 blocks per level,
and 3/4 of the dimensions is splitted out at each level.

\paragraph{Optimization details.} We use Adamax optimizer~\citep{kingma2014adam} with default $\beta_1$ and $\beta_2$. And learning
rate is set to 0.01 and exponentially decreases to 0.001 with $1\times 10^3$ decay steps and 0.98 decay rate.
We use the gradient checkpointing trick~\citep{chen2016training} to improve the memory utilization.
Batch size is set to 1024 for CIFAR10 and ImageNet 32$\times$32, and 4 for CelebA-HQ 1024$\times$1024.
The preprocessing and division of datasets follow the method used in~\citet{kingma2018glow}.

\begin{figure}[t]
  \centering
  \setlength{\tabcolsep}{1.5pt}
  \begin{tabular}{ccc}
    \raisebox{0.01\height}{\includegraphics[width=0.102\textwidth]{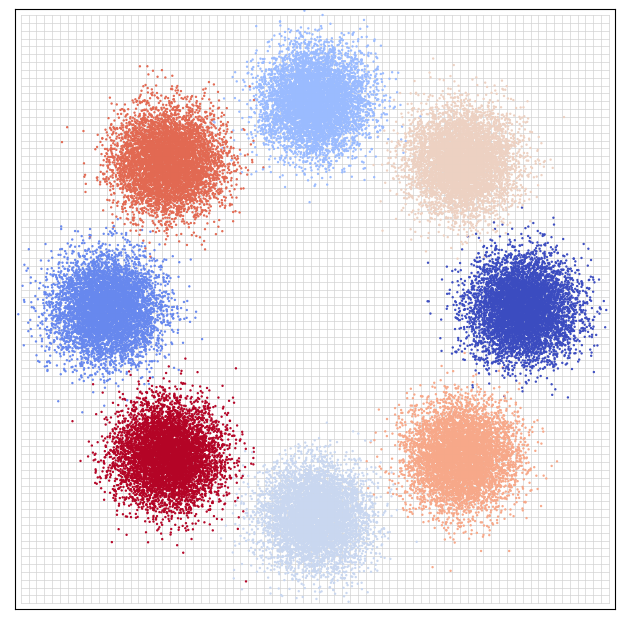}} &
    \includegraphics[width=0.6\textwidth]{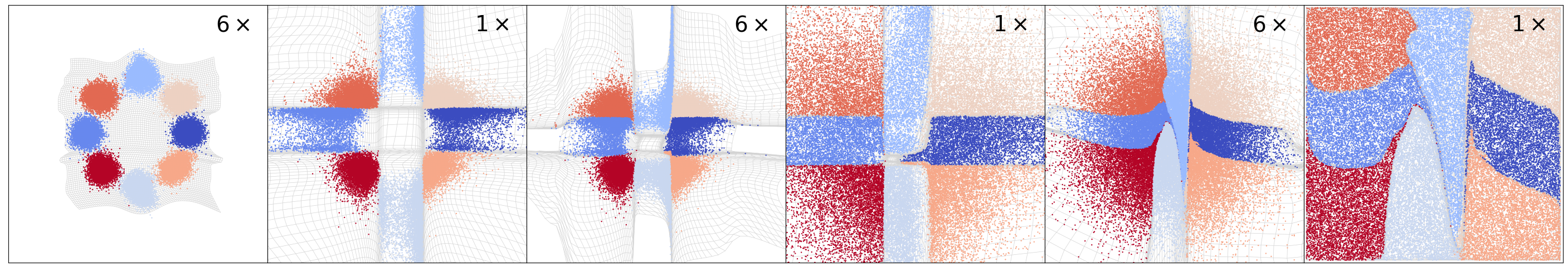} &
    \raisebox{0.02\height}{\includegraphics[width=0.102\textwidth]{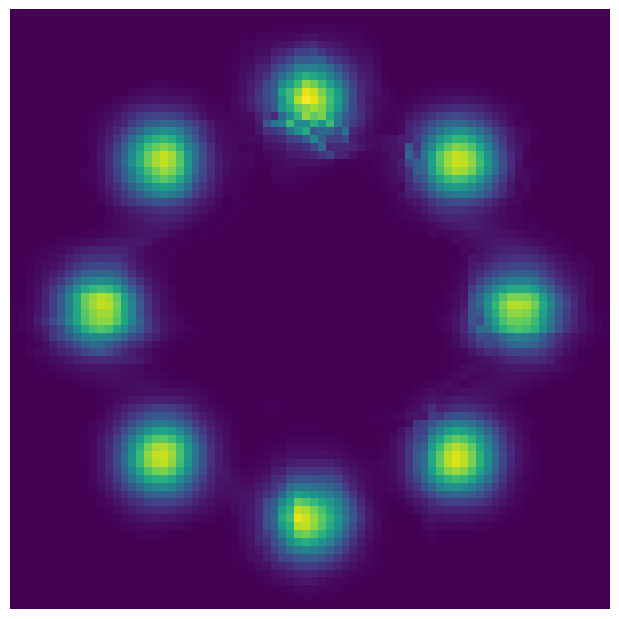}} \\
    \raisebox{0.01\height}{\includegraphics[width=0.102\textwidth]{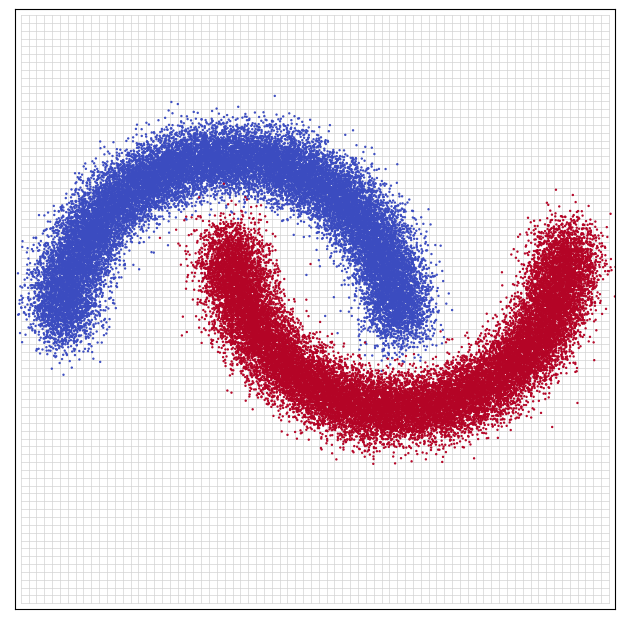}} &
    \includegraphics[width=0.6\textwidth]{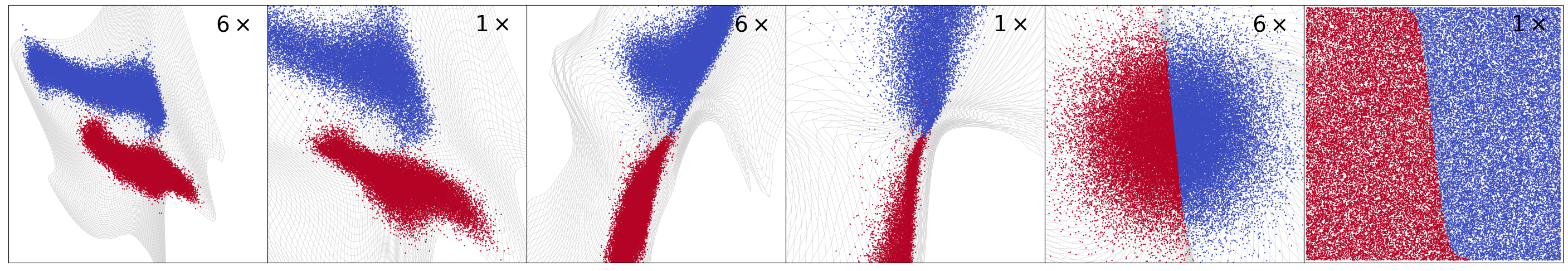} &
    \raisebox{0.02\height}{\includegraphics[width=0.102\textwidth]{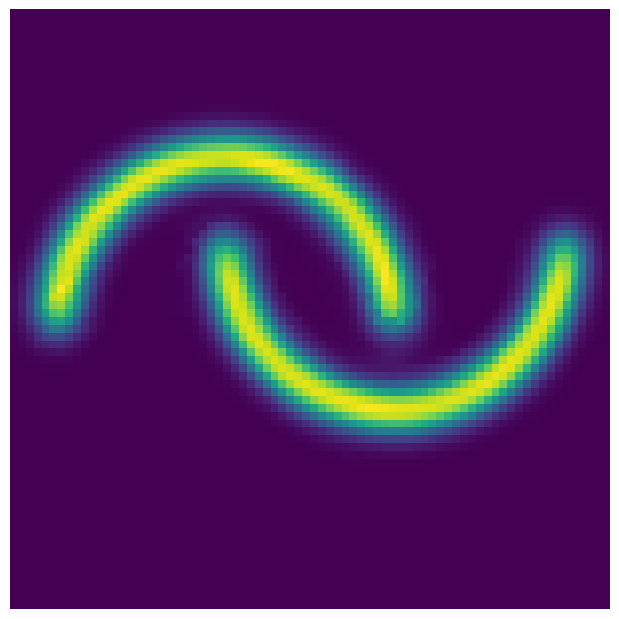}} \\
    \raisebox{0.01\height}{\includegraphics[width=0.102\textwidth]{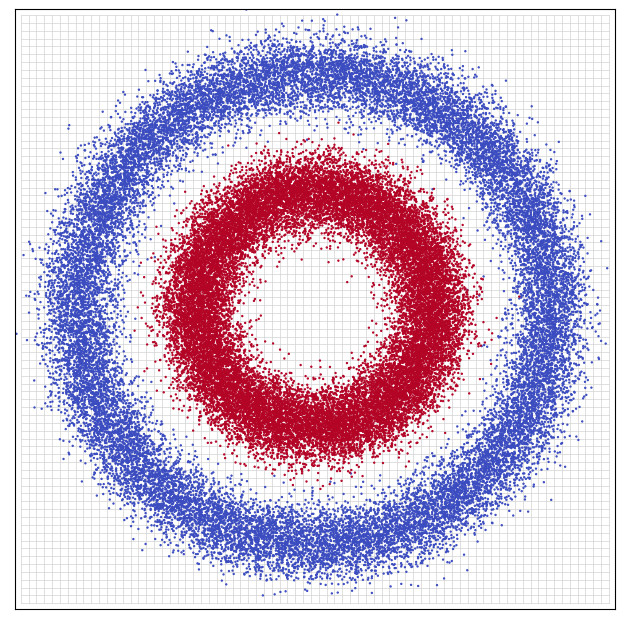}} &
    \includegraphics[width=0.6\textwidth]{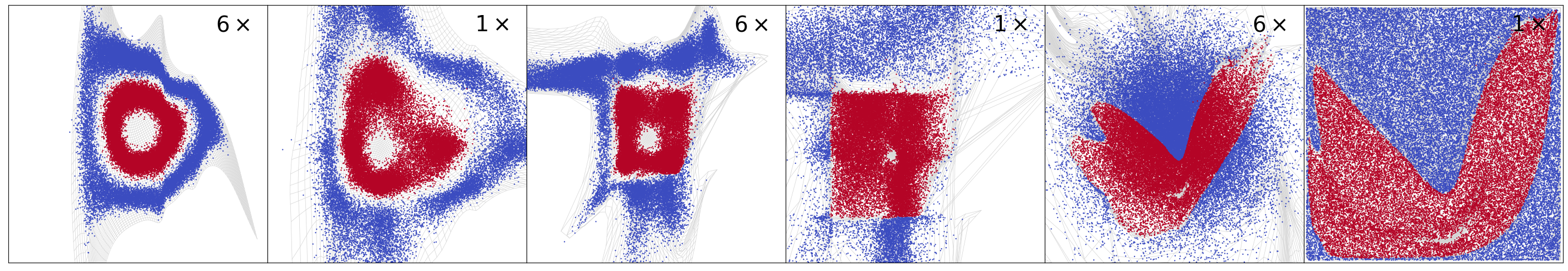} &
    \raisebox{0.02\height}{\includegraphics[width=0.102\textwidth]{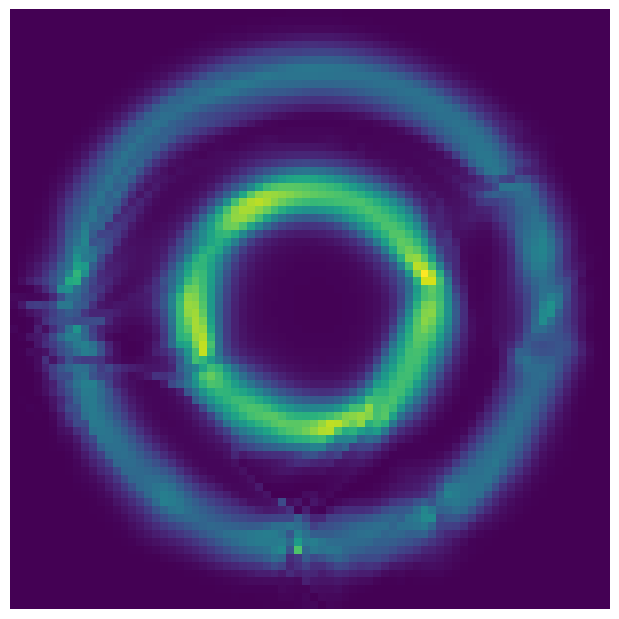}} \\
  \end{tabular}
  \caption{Illustration of diffeomorphic deformations on 2D toy problems. \textbf{Leftmost:} Samples from empirical distribution.
  \textbf{Middle:} Data distribution of intermediate mappings in the order of layer number. The symbol "$N\times$" in each grid denotes
  that the displayed domain is $N$ times of the leftmost column. \textbf{Rightmost:} Density estimation by flow. The intermediate mappings
  reshape the data distribution by iteratively performing non-uniform stretching (expansive mapping) and compression
  (contractive mapping) on its domain.}
  \label{fig:extra_2dtoy}
\end{figure}

\begin{figure}[b]
  \centering
  \setlength{\tabcolsep}{5pt}
  \begin{tabular}{cc}
  \includegraphics[width=0.22\textwidth]{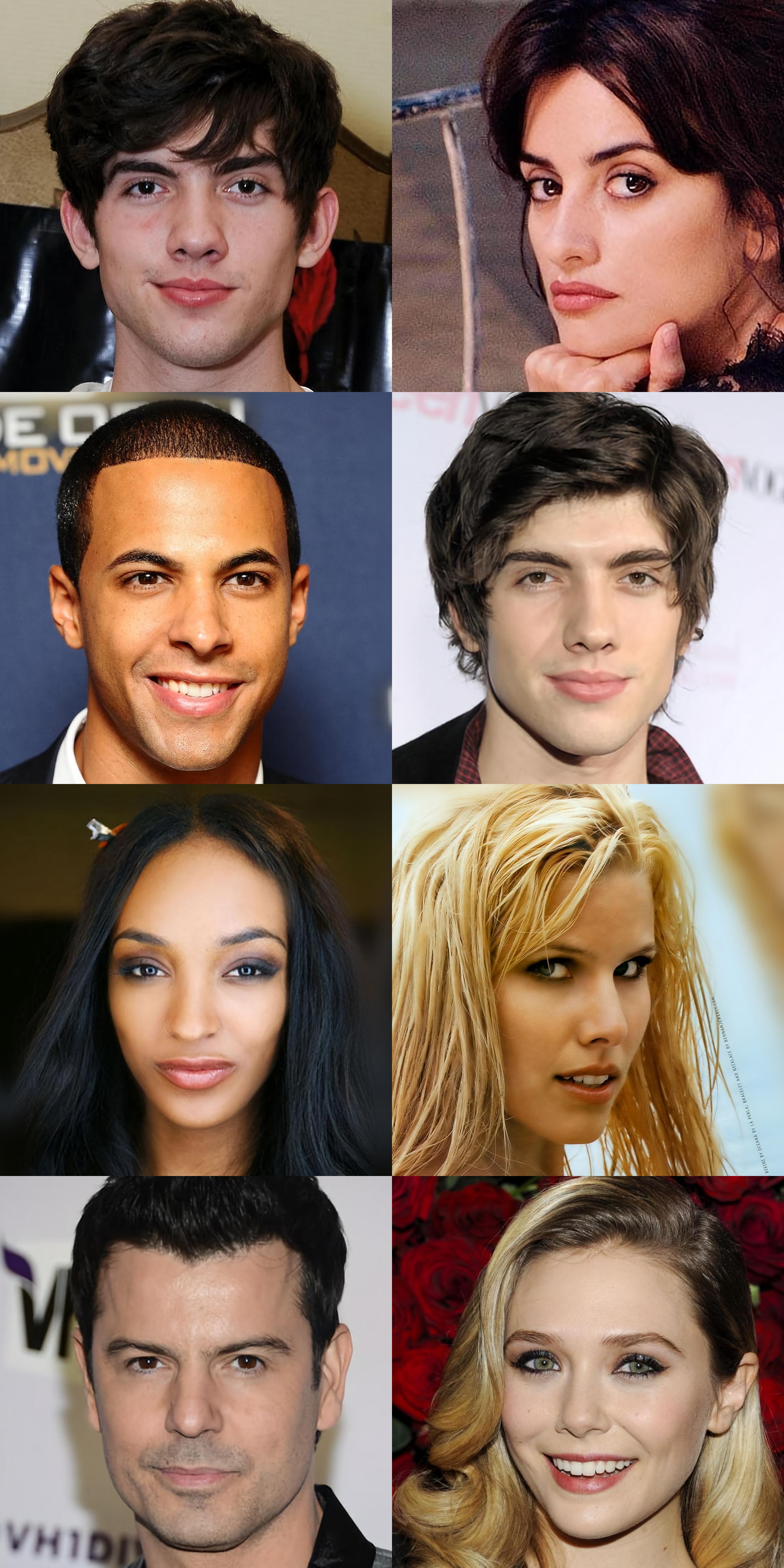} &
  \includegraphics[width=0.66\textwidth]{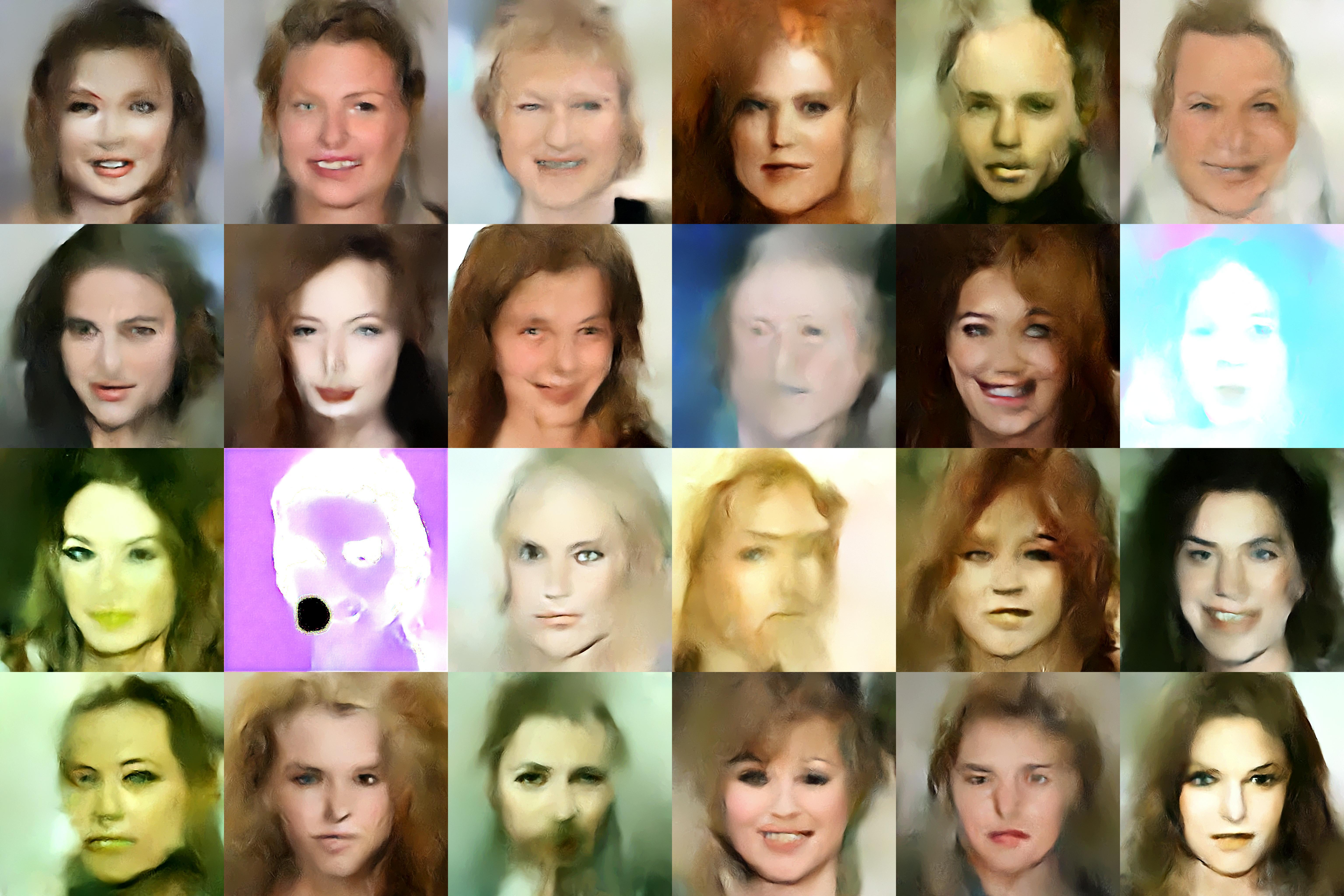}
  \end{tabular}
  \caption{\textbf{Left:} Real samples. \textbf{Right:} Randomly generated samples from 5bit CelebA-HQ 1024$\times$1024, with temperature 0.9.}
  \label{fig:celeba1024}
\end{figure}

\section{Extra Samples}
Additional samples on 2D toy problems and generated samples from CelebA-HQ 1024$\times$1024 are shown in Fig.~\ref{fig:extra_2dtoy} and
Fig.~\ref{fig:celeba1024}, respectively.


\end{document}